\documentclass[acmsmall,screen, nonacm]{acmart}    % REMOVE noacm

\AtBeginDocument{%
  }

\usepackage{graphicx}
\usepackage{subcaption}
\usepackage{pdflscape}
\usepackage{multirow}
\usepackage{comment}
\usepackage{algorithm} 
\usepackage{algpseudocode}
\usepackage{adjustbox}

\begin{document}

\title[A Performance Analysis of Selection Methods in GP for Symbolic Regression]{A Performance Analysis of Lexicase-Based and Traditional Selection Methods in GP for Symbolic Regression}

\author{Alina Geiger}
\email{geiger@uni-mainz.de}
\orcid{0009-0002-3413-283X}
\affiliation{
  \institution{Johannes Gutenberg University Mainz}
  %\city{Mainz}
  \country{Germany}
}

\author{Dominik Sobania}
\email{dsobania@uni-mainz.de}
\orcid{0000-0001-8873-7143}
\affiliation{
  \institution{Johannes Gutenberg University Mainz}
  %\city{Mainz}
  \country{Germany}
}

\author{Franz Rothlauf}
\email{rothlauf@uni-mainz.de}
\orcid{0000-0003-3376-427X}
\affiliation{
  \institution{Johannes Gutenberg University Mainz}
  %\city{Mainz}
  \country{Germany}
}

\renewcommand{\shortauthors}{Geiger et al.}

\acmArticleType{Research}

\acmCodeLink{}
\acmDataLink{}

%\acmContributions{AG, DS, and FR designed the study; AG conducted the experiments and analyzed the results; all authors participated in writing the manuscript.}

\keywords{Symbolic Regression, Genetic Programming, Lexicase Selection, Down-sampling}

\begin{abstract}

   In recent years, several new lexicase-based selection variants have emerged due to the success of standard lexicase selection in various application domains. For symbolic regression problems, variants that use an $\epsilon$-threshold or batches of training cases, among others, have led to performance improvements.  
   Lately, especially variants that combine lexicase selection and down-sampling strategies have received a lot of attention. This paper evaluates the most relevant lexicase-based selection methods as well as traditional selection methods in combination with different down-sampling strategies on a wide range of symbolic regression problems. In contrast to most work, we not only compare the methods over a given evaluation budget, but also over a given time budget as time is usually limited in practice.
   We find that for a given evaluation budget, $\epsilon$-lexicase selection in combination with a down-sampling strategy outperforms all other methods. If the given running time is very short, lexicase variants using batches of training cases perform best. Further, we find that the combination of tournament selection with informed down-sampling performs well in all studied settings. 
\end{abstract}

\maketitle

\section{Introduction}\label{sec:intro}

Symbolic regression searches for a  mathematical expression that best describes a given problem~\cite{Poli.2008}. Applications of symbolic regression range from finance~\cite{Chen.2012}, and medicine~\cite{la2023flexible} to materials science~\cite{hernandez2019fast}. Genetic Programming (GP)~\cite{Koza.1992} is an evolutionary computation technique that has often been used successfully to solve even complex symbolic regression tasks \cite{LaCava.2021}. In GP, a population of individuals evolves in an evolutionary process guided by selection and modified through variation operators. Recent work has shown that especially the choice of the selection method has a major impact on the solution quality~\cite{Helmuth.07082020}. 

Traditional selection methods like tournament selection or fitness-proportionate selection evaluate the quality of individuals based on an aggregated fitness score leading to a loss of information about the structure of the training data~\cite{Krawiec.2014}. Therefore, lexicase selection~\cite{Spector.2012, Helmuth.2014} has been proposed that uses the error values of individuals on each training case. 
This allows lexicase selection to select specialists that perform particularly well for certain parts of the problem, which has been found to be beneficial for search towards better solutions~\cite{Helmuth.2019, Helmuth.2020b}. 

Due to the successful application of lexicase selection in many problem domains~\cite{Helmuth.2015, Moore.2017, LaCava.2016, Aenugu.2019}, several variants of lexicase selection have been proposed. In the domain of symbolic regression, $\epsilon$-lexicase selection outperformed standard lexicase selection and tournament selection on the considered problems~\cite{LaCava.2016, LaCava.2019}.  Others suggested to use batches of training cases instead of single training cases to evaluate the quality of individuals \cite{de2019batch, Aenugu.2019, Sobania.2022}. 

Recently, especially variants combining lexicase selection with down-sampling strategies received much  attention because it led to higher problem-solving success in different applications \cite{boldi2024informed, geiger.2023, Hernandez.2019, Ferguson.2020}. Down-sampling uses only a subset of the training cases  in each generation to evaluate the quality of the individuals. For a fixed evaluation budget, the saved evaluations can be allocated to a longer search which allows the exploration of more individuals. The simplest down-sampling strategy is to randomly sample a percentage of the training cases in each generation~\cite{Ferguson.2020, Hernandez.2019}. However, this might cause the exclusion of important training cases for several generations~\cite{boldi2024informed}. Therefore, ~\citeN{boldi2024informed} proposed informed down-sampling as a strategy to create more diverse subsets. 

In our previous conference paper~\cite{geiger2024comprehensive}, we analyzed and compared several relevant lexicase-based selection methods in combination with random down-sampling. We used tournament selection as baseline and also studied the combination of lexicase variants with batches of training cases. 
The paper at hand builds upon this work and presents a more comprehensive and more detailed comparison of lexicase-based selection methods. In addition to tournament selection, we include fitness-proportionate selection as a baseline. \citet{geiger2024comprehensive} analyzed selection methods only in combination with random down-sampling. The paper at hand also adds selection variants using informed down-sampling.
We compare the methods not only for a given evaluation budget (as is common in literature) but also for a given time budget as the selection methods differ in terms of time complexity. For the comparison, we extend the problem set to 26 synthetic and real-world benchmark problems. Our goal is to provide researchers as well as practitioners with a comprehensive guide for choosing the appropriate selection method.
 
We find that $\epsilon$-lexicase selection in combination with random or informed down-sampling performs best for a given evaluation budget. Further, we observe that the relative performance of each selection method depends on the given running time. For example, for a time budget of 24h, the best performing methods are tournament selection in combination with informed down-sampling and $\epsilon$-lexicase selection in combination with random down-sampling. However, if the running time is very short (1h or less), batch-tournament selection and batch-$\epsilon$-lexicase selection perform best. Additionally, the detailed analysis for each problem shows that while there are general trends, the relative performance of the methods varies between problems. 

Sect.~\ref{sec:related_work} provides a brief overview of the related work. In Sect.~\ref{sec:selection_methods}, we describe the selection methods analyzed in this study, followed by a description of the applied down-sampling strategies in Sect.~\ref{sec:down_sampling}. Section~\ref{sec:experimental_methods} presents our experimental setting and the benchmark problems. In Sect.~\ref{sec:results} we present our results, followed by the conclusions in Sect.~\ref{sec:conclusions}.

\section{Related Work}\label{sec:related_work}

We briefly discuss prior work on lexicase selection. More detailed descriptions of the lexicase-based selection methods and down-sampling strategies are given in Sec.~\ref{sec:selection_methods} and~\ref{sec:down_sampling}.

Traditional selection methods (e.g., tournament selection and fitness-proportionate selection) evaluate the quality of individuals based on an aggregated fitness value. This  leads to a loss of information about the structure of the training data \cite{Krawiec.2014}. Lexicase selection~\cite{Spector.2012, Helmuth.2014} 
has been proposed as an alterantive considering the errors of individuals on each training case separately. Lexicase selection is able to select specialists as parents, meaning individuals that solve some part of a problem better than others while performing worse on average~\cite{Helmuth.2020b, Pantridge.2018, Helmuth.2019}. In addition, prior work has found that lexicase selection maintains a higher population diversity compared to tournament selection~\cite{Helmuth.2016, Helmuth.2016c}. 

For continuous problems, standard lexicase selection is not able to achieve the same performance improvements due to the selection of strictly elite individuals on training cases~\cite{LaCava.2016}. 
Therefore, lexicase variants with a relaxed pass condition have been proposed~\cite{LaCava.2016, Spector.2018}, with $\epsilon$-lexicase selection~\cite{LaCava.2016, LaCava.2019} being the best known. 
Additionally, variants using batches of training cases have been introduced~\cite{Aenugu.2019, Sobania.2022, de2019batch}. It has been found that the use of batches can increase the generalizability of the found solution~\cite{Aenugu.2019, Sobania.2022}. Batch-tournament selection ~\cite{de2019batch} uses batches in combination with tournament selection to improve the efficiency of lexicase selection. 
Another idea to improve the runtime of lexicase selection is to combine lexicase selection and weighted shuffle with partial evaluation ~\cite{ding2022lexicase, ding2022going}. Plexicase selection~\cite{ding2023probabilistic} improves the runtime of lexicase selection by sampling individuals from a probability distribution instead of doing the actual selection procedure. \citeN{ni2024dalex} proposed DALex to improve the runtime by performing all calculations as matrix multiplications. In other studies, the combination of lexicase selection and novelty search has been explored to prevent premature convergence ~\cite{Kelly.2019, jundt2019comparing}.

Recently, the combination of lexicase variants with down-sampling strategies has been found to lead to higher performance~\cite{boldi2024informed, geiger.2023, Hernandez.2019, Ferguson.2020}. Therefore, the influence of down-sampling has been analyzed in several studies~\cite{Helmuth.2020c, Helmuth.2021, Schweim.2022, Hernandez.2021, Boldi2023.static}. Down-sampling reduces the number of evaluations per generations by sampling a subset of training cases from the training set allowing the evaluation of more individuals with the same evaluation budget~\cite{Helmuth.2020c}. While random down-sampling~\cite{Hernandez.2019, Ferguson.2020} randomly creates subsets, informed down-sampling~\cite{boldi2024informed} creates subsets that includes distinct training cases. It has been found that informed down-sampling outperforms random down-sampling in combination with lexicase selection on program synthesis problems~\cite{boldi2024informed}. Therefore, informed down-sampling has been further studied in combination with tournament and fitness-proportionate selection for program synthesis problems and synthetic regression problems with an exact solution~\cite{boldi2024untangling, Boldi2023.problemsolvingbenefits}. However, to our knowledge, there is no study that analyzes the influence of informed down-sampling for continuous symbolic regression problems.

Lexicase selection and its variants have been successfully applied and analyzed in several domains, like program synthesis~\cite{Helmuth.07082020, Helmuth.2015, Sobania.2023, Sobania.06262021}, symbolic regression~\cite{LaCava.2016, LaCava.2019, geiger.2023}, evolutionary robotics~\cite{Moore.2017, Moore.2018}, rule-based learning systems~\cite{Aenugu.2019, Wagner.2021}, and deep learning~\cite{ding2021optimizing}.

Prior work analyzed and compared several lexicase variants for program synthesis problems~\cite{Helmuth.07082020}. However, to our knowledge, a study that compares lexicase-based selection methods in combination with different down-sampling strategies, including informed down-sampling, on a wide range of symbolic regression problems is missing so far. 

\section{Selection Methods for Symbolic Regression Problems}\label{sec:selection_methods}

We provide a detailed description of tournament selection, fitness-proportionate selection, lexicase selection, as well as the lexicase-based variants studied in this work in the context of symbolic regression. 

\subsection{Tournament Selection}

With tournament selection, $k$ individuals are randomly chosen to participate in a tournament. The participant with the best fitness wins the tournament and is selected as a parent \cite{Poli.2008}. Fitness is measured as an aggregated value, for example the mean squared error (MSE)
\begin{equation}\label{MSE}
\mathrm{MSE}(T) = \frac {1}{|T|} \sum_{t \in T} (y_t - \hat{y}_t)^2,
\end{equation}
where $\hat{y}_t$ is the predicted output of an individual and $y_t$ the desired output for all training cases $t \in T$.

\subsection{Fitness-Proportionate Selection}
With fitness-proportionate selection (also known as roulette wheel selection) an individual $i$ is selected as a parent with the probability of

\begin{equation}\label{fps}
p_i = \frac {f_i}{\sum_{j=1}^{N} f_j},
\end{equation}
where $f_i$ is the fitness of individual $i$ and $N$ is the number of individuals in the population~\cite{banzhaf1998genetic}. This means individuals with higher fitness have a higher probability of being selected than individuals with  lower fitness. As with tournament selection, fitness is measured in terms of an aggregated value. 

\subsection{Lexicase Selection}

In contrast to tournament selection and fitness-proportionate selection, lexicase selection~\cite{Spector.2012} considers the error values $e_t$ on each training case $t \in T$ separately rather than comparing individuals based on an aggregated fitness value. Algorithm~\ref{algo:lexicase} shows the procedure of a single selection event with lexicase selection with population $P$, the set of training cases $T$, the error $e_t$ of an individual $i \in P$ for training case $t\in T$, and the minimal error $e^*_t$ on cases $t\in T$.

\begin{algorithm}[H]
	\caption{Lexicase Selection}
	\label{algo:lexicase}
	\begin{algorithmic}[1]
    \State $C$ := all individuals $i \in P$\label{line:init_candidates}
    \State $T'$ := randomly shuffled $T$\label{line:init_cases}
        \While {$|T'|$ $>$ 0 and $|C|$ $>$ 1}\label{line:loop_sel}
            \State $t$ := first case of $T'$
            \State $C$ := all candidates with $e_t = e^*_t$ \label{line:pass_condition}
            \State \textbf{remove} $t$ from $T'$
        \EndWhile\label{line:end_loop_sel}
\State  \textbf{return} random choice from $C$\label{line:return_offspring}
	\end{algorithmic} 
\end{algorithm}

With lexicase selection, all individuals $i \in P$ are considered as candidates for selection (line~\ref{line:init_candidates}) and all training cases $t \in T$ are randomly ordered (line~\ref{line:init_cases}). Lexicase selection iterates through all training cases always keeping only the individuals in the candidate pool $C$ with the lowest error $e^*_t$ on the current case $t$ (line~\ref{line:pass_condition}). This is repeated until there is only one candidate left or there are no more training cases. In this case, one individual is randomly chosen from the remaining candidate pool $C$ (line~\ref{line:return_offspring}) \cite{Spector.2012, Helmuth.2014}.

\subsection{$\epsilon$-Lexicase Selection}
For continuous problems, the pass condition of standard lexicase selection that only candidates with the lowest error remain in the candidate pool (Algorithm~\ref{algo:lexicase}, line~\ref{line:pass_condition}) is too strict. Therefore, $\epsilon$-lexicase selection has been proposed by \citeN{LaCava.2016, LaCava.2019} for solving symbolic regression problems. 

$\epsilon$-lexicase selection modifies the pass condition of standard lexicase selection in such a way that all candidates with an error $e_t <= e^*_t + \epsilon_t$ remain in the candidate pool.
$\epsilon_t$ is calculated for a training case $t$ using the median absolute deviation \cite{PhamGia.2001} 
\begin{equation}\label{MAD}
\epsilon_t = \mathrm{median}(|\textbf{e}_t - \mathrm{median}(\textbf{e}_t)|),
\end{equation}
where $\textbf{e}_t$ is a vector of all errors on the current case $t$ across the candidate pool $C$~\cite{LaCava.2019}.

\subsection{$\epsilon$-Plexicase Selection}
\citeN{ding2023probabilistic} proposed plexicase selection as a more efficient alternative to lexicase selection. The idea is to reduce the runtime of lexicase selection by sampling individuals from a probability distribution instead of doing the actual selections. However, \citeN{dolson2023calculating} has shown that the problem of calculating the actual selection probabilities using lexicase selection is $\mathcal{NP}$-hard. Therefore, \citeN{ding2023probabilistic} only approximate the probability of individuals being selected with lexicase selection by finding the Pareto set boundaries through pairwise comparisons of individuals and assigning selection probabilities to each individual based on these comparisons. In order to further improve the solution quality, \citeN{ding2023probabilistic} introduce a hyperparameter $\alpha$ to manipulate the generated probability distribution.
For symbolic regression problems, an $\epsilon$-threshold like in $\epsilon$-lexicase is used to calculate $\epsilon$-relaxed Pareto set boundaries. They call this variant $\epsilon$-plexicase selection.

\subsection{Batch-Tournament Selection}
The problem-solving performance of lexicase selection is superior to tournament selection \cite{Spector.2012, Helmuth.2014}. However, lexicase selection is computationally expensive compared to tournament selection~\cite{LaCava.2019}. Therefore, \citeN{de2019batch} proposed batch-tournament selection as well as the variant Batch Tournament Selection Shuffled (BTSS) which shuffles the cases in addition to grouping them in batches. Both approaches combine the good performance of lexicase selection with the low computational effort of tournament selection. 

For BTSS, the training cases are randomly combined to batches of size $b$ and the fitness values of the individuals are derived from their performance on each batch (e.g., in terms of the MSE). For each selection event, $k$ individuals participate in a tournament. The individual with the lowest MSE on the current batch is selected as a parent. For each selection event, another batch is used to compare the quality of the individuals. If there are less batches than parent selection events, the batches are used multiple times.

\subsection{Batch-$\epsilon$-Lexicase Selection}
Using lexicase selection, the outcome of a selection event highly depends on the ordering of training cases~\cite{Aenugu.2019}. Therefore, lexicase variants using batches of training cases instead of single training cases to compare the performance of individuals in the lexicase selection process have been proposed~\cite{Aenugu.2019, Sobania.2022}. It has been found that the generalization ability of the found solutions is better using batch-lexicase selection compared to standard lexicase selection~\cite{Aenugu.2019, Sobania.2022}.
For continuous-valued problems,~\citeN{geiger2024comprehensive} proposed Batch-$\epsilon$-lexicase selection to combine batches with $\epsilon$-lexicase selection. In each generation, the training cases are randomly combined to batches of size $b$ and the fitness values of the individuals are calculated on each batch (e.g., in terms of the MSE). The selection is then performed with $\epsilon$-lexicase selection. 

\section{Down-sampling Strategies}\label{sec:down_sampling}
We describe the down-sampling strategies random down-sampling and informed down-sampling. 

\subsection{Random Down-sampling}
In  application domains, like program synthesis \cite{Ferguson.2020, Hernandez.2019, Helmuth.2021} and symbolic regression \cite{geiger.2023}, it has been found that the combination of lexicase-based selection methods and random down-sampling \cite{Goncalves.2012} improves the problem-solving success. Random down-sampling means that in each generation a random subset of training cases is used to evaluate the performance of the individuals. Therefore, the saved fitness evaluations in each generation can be used to  search for more generations or to increase the population size. This is beneficial as more individuals can be explored with the same evaluation budget \cite{Helmuth.2020c}. Since a different random subset of training cases is used in each generation, the population is likely to be evaluated on a large proportion of the training cases over a few generations \cite{Hernandez.2019}. 

The down-sampling rate is defined by a parameter $d$. For example, if $d = 0.1$, only $10\%$ of the training cases are used in each generation, meaning the search can be performed 10 times longer or the population size can be increased by a factor of $10$.

\subsection{Informed Down-sampling}
The random creation of subsets of training cases has the problem that important cases might be excluded for several generations while training cases with redundant information might end up in the same subset \cite{Ferguson.2020, boldi2024informed}. Therefore, \citeN{boldi2024informed} proposed informed down-sampling to create subsets consisting of more diverse training cases. The authors define synonymous cases as cases that are solved by the same individuals in a population. If two cases are solved by different individuals, this indicates that the cases measure a different functionality. Hence, it would be beneficial to have both cases in a subset.

\citeN{boldi2024informed} define the distance between two cases as the Hamming distance between their two solve vectors (a solve vector contains the output of each individual for that case). Then, they use the Farthest First Traversal Algorithm \cite{hochbaum1985best} to select a subset of training cases with a large pairwise distance. 
However, evaluating all individuals on all training cases in each generation to create informed subsets would mean that the benefit of down-sampling is lost because there are no saved fitness evaluations. Therefore, \citeN{boldi2024informed} propose to sample a fraction $s$ of the parents that are evaluated on all training cases in order to calculate the distance between training cases. In addition, the distance calculation is only performed every $g$ generations to further reduce the number of fitness evaluations required to create the informed subsets. 

Since \citeN{boldi2024informed} study informed down-sampling for program synthesis problems, solve vectors are binary. In the domain of symbolic regression, the solve vectors contain continuous errors values. Therefore, we calculate the distances between two cases in this work using Euclidean distances instead of  Hamming distances.

\section{Experimental Setting}\label{sec:experimental_methods}

We present our experimental setup, followed by a description of the used benchmark problems.

\subsection{Setup}\label{sec:experimental_setting}
All experiments were implemented within the DEAP framework~\cite{Fortin.2012} (version 1.4.1). Table~\ref{tab:parameter_setting} summarizes the parameter values of our GP setting which is in line with previous work~\cite{geiger.2023, geiger2024comprehensive}.  

\begin{table}[h]
  \centering
\caption{Parameter setting of our GP approach.}
\label{tab:parameter_setting}
\begin{tabular}{l|r}
 \toprule 
\textbf{Parameter} & \textbf{Value} \\
\midrule
Population size & $500$ \\
Primitive set & $\{\textrm{\textbf{x}}, \textrm{ERC}, +, -, *, \textrm{AQ}, \textrm{sin}, \textrm{cos}, \textrm{neg}\}$ \\
ERC values & $\{-1,0,1\}$ \\
Initialization method & Ramped half-and-half \\
Maximum tree depth & $17$ \\
Crossover probability & $80\%$ \\
Mutation probability & $5\%$ \\
Runs & 30 \\
\bottomrule
\end{tabular}
\end{table}

The primitive set consists of all input features $\textbf{x}$, an ephemeral random constant (ERC) where $\mathrm{ERC} \in \{-1,0,1\}$, and the arithmetic functions addition, subtraction, multiplication, analytic quotient (AQ) \cite{Ni.2013}, sine, cosine, and negative. 
The population is initialized using ramped half-and-half with tree depths between $0$ and $4$ and a population size of $500$. In each generation, crossover is applied with a probability of $80\%$ and mutation with a probability of $5\%$. We restrict the maximum tree depth to $17$~\cite{Koza.1992}. Each configuration is repeated 30 times. 

Usually, selection methods are compared for a given evaluation budget which is determined by the number of training cases, the population size and the number of generations \cite{geiger.2023, Helmuth.2020c}. If down-sampling is applied, we allocate the saved evaluations to a longer search. Table~\ref{tab:generational_limit} shows the number of generations given to each run depending on the down-sampling strategy applied. 

\begin{table}[h]
  \centering
\caption{Generational limit for runs without down-sampling, with random down-sampling, and with informed down-sampling.}
\label{tab:generational_limit}
\begin{tabular}{l|r}
 \toprule 
\textbf{Down-sampling Strategy} & \textbf{\# Generations} \\
\midrule
No down-sampling ($d=1.0$) & $100$ \\
Random down-sampling ($d=0.1$) & $1,000$ \\
Informed down-sampling ($d=0.1, s=0.01, g=10$) & $991$ \\
\bottomrule
\end{tabular}
\end{table}

If there is no down-sampling used ($d = 1.0$), we set the number of generations to $G = 100$. For random down-sampling we set a down-sampling rate of $d = 0.1$ as suggested by \citeN{geiger.2023}, meaning we can search for $\hat{G} = 1,000$ generations with the same evaluation budget. For runs with informed down-sampling, the down-sampling rate is set to $d = 0.1$ as well. The parent sampling rate is set to $s = 0.01$ and the distance calculation scheduling parameter is set to $g = 10$ as suggested by \citeN{boldi2024informed}. This means that every 10 generations, 5 individuals are evaluated on all training cases to calculate the distance between the cases. The generational limit for informed down-sampling is calculated as 
\begin{equation}
\hat{G} = \frac{G}{d+\frac{s(1-d)}{g}}
\end{equation}
where $G$ is the number of generations for runs without down-sampling, $d$ is the down-sampling rate, $s$ the parent sampling rate and $g$ the distance calculation scheduling parameter \cite{boldi2024informed}. In our case, this leads to limit of $\hat{G} = 991$ for runs using informed down-sampling. 

For each problem, we randomly split all cases into $70\%$ training cases, $15\%$ validation cases, and $15\%$ test cases. The training cases are used to evaluate the fitness of individuals during the evolutionary process to perform the selection. In contrast, the validation cases are used to select a final solution. Since the validation cases were not used for the fitness calculation, we avoid selecting a final solution that is overfitted to the training cases. In detail, the individual with the lowest MSE on the validation cases is stored as the current best solution in the Hall of Fame in each generation.\footnote{Due to the use of a validation fitness, individuals in the Hall of Fame are compared based on an aggregated fitness value. This ensures that the final solution performs well overall.} After a run, the final solution is evaluated on the unseen test cases in terms of the MSE. 

The calculation of the fitness of individuals during the search depends on the selection method. As tournament selection and fitness-proportionate selection evaluate individuals based on an aggregated value, we use the MSE as the fitness measure. For fitness-proportionate selection, the selection probabilities of individuals are assigned based on the inverse of their fitness (so individuals with a lower MSE have a higher probability of being selected).\footnote{In case an individual has a fitness of zero, $\epsilon = 1e-10$ is added to the error values in the population to avoid zero divisions. This only happened during a few runs for problem 523\_analcatdata\_neavote.} 
For $\epsilon$-lexicase selection and $\epsilon$-plexicase selection it is necessary to evaluate the performance of individuals on each training case separately. Therefore, we calculate for each individual the squared error on each training case. For batch-tournament selection and batch-$\epsilon$-lexicase selection, the performance of individuals is measured in terms of the MSE on each batch of training cases. 

The parameter settings of the selection methods are shown in Table~\ref{tab:parameter_selection}. They are carefully chosen according to the recommendations from the literature. 

\begin{table}[h]
  \centering
\caption{Parameter settings of the selection methods.}
\label{tab:parameter_selection}
\begin{tabular}{l|r|r}
 \toprule 
\textbf{Selection Method} & \textbf{Parameter} & \textbf{Value} \\
\midrule
Tournament selection & tournament size $k$ & 5 \\
$\epsilon$-Plexicase selection & alpha $\alpha$ & 1.0 \\
Batch-$\epsilon$-lexicase selection & batch size $b$ & $\{0.05, 0.075, 0.1\}$ \\
\multirow[t]{2}{*}{Batch-Tournament selection} & tournament size $k$ & 64 \\
 & batch size $b$ & $\{0.05, 0.075, 0.1\}$ \\
\bottomrule
\end{tabular}
\end{table}

For tournament selection, we set a tournament size of $k = 5$ \cite{Fang.2010}. The parameter of $\epsilon$-plexicase selection is set to $\alpha = 1.0$ as suggested by \citeN{ding2023probabilistic}. 
For batch-$\epsilon$-lexicase selection, we study batch sizes $b \in \{0.05, 0.075, 0.1\}$~\cite{Aenugu.2019}, which means that a single batch consists of $5\%$, $7.5\%$ or $10\%$ of the training cases. 
For batch-tournament selection, we set a tournament size of $k = 64$~\cite{de2019batch} and we study batch sizes $b \in \{0.05, 0.075, 0.1\}$.

In addition to runs given a fixed evaluation budget, we also study the performance for a given running time as recommended in literature~\cite{ravber2022maximum}. We measure time in terms of wall clock time. In each generation, we track the performance of the current best individual on the test cases, the median tree size across the population and the time. We conducted the experiments on a high performance computing cluster using Intel 2630v4 2,20GHz CPUs. All processes were executed single threaded. Please note that the comparison of methods over time always depends on the implementation. Here, all methods have been implemented within the same framework and all experiments run on the same hardware to minimize this limitation.

\subsection{Benchmark Problems}\label{sec:problems}

We study 26 problems taken from SR Bench/PMLB~\cite{LaCava.2021, romano2022pmlb} with the number of instances ranging from $100$ to $1,059$ and the number of features ranging from $2$ to $124$.\footnote{Due to limited computational resources, we only considered problems with a maximum of around $1,000$ instances.} Table~\ref{tab:problems} shows all considered problems with the number of instances, the number of features, and the type (real-world or synthetic). We extended the problem set from our previous paper~\cite{geiger2024comprehensive} by 6 real-world problems to have the same number of real-world and synthetic problems and to cover a wide range of applications and problem sizes.

\begin{table}[h]
    \centering
    \caption{All benchmark problems used to evaluate the selection methods with the number of instances, the number of features and the problem type.}
    \label{tab:problems}
    \begin{tabular}{l|r|r|r}
    \toprule
Problem & \# Instances & \# Features & Type \\
\midrule
1027\_ESL                 & 488   & 4 & real-world\\
1028\_SWD                 & 1000  & 10 & real-world\\ 
1030\_ERA                 & 1,000  & 4 & real-world\\
207\_autoPrice            & 159   & 15 & real-world\\
230\_machine\_cpu         & 209   & 6 & real-world\\
4544\_GeographicalOriginalofMusic & 1,059 &  117  & real-world\\
505\_tecator              & 240       &  124  & real-world\\
519\_vinnie               & 380   & 2 & real-world\\
522\_pm10                 & 500   & 7 & real-world\\
523\_analcatdata\_neavote & 100   & 2 & real-world\\
560\_bodyfat              & 252   & 14 & real-world\\
581\_fri\_c3\_500\_25     & 500   & 25 & synthetic\\
589\_fri\_c2\_1000\_25    & 1,000  & 25 & synthetic\\
591\_fri\_c1\_100\_10     & 100   & 10 & synthetic\\
606\_fri\_c2\_1000\_10    & 1,000  & 10 & synthetic\\
607\_fri\_c4\_1000\_50    & 1,000  & 50 & synthetic\\
615\_fri\_c4\_250\_10     & 250   & 10 & synthetic\\
617\_fri\_c3\_500\_5      & 500   & 5 & synthetic\\
621\_fri\_c0\_100\_10     & 100   & 10 & synthetic\\
623\_fri\_c4\_1000\_10    & 1,000  & 10 & synthetic\\
624\_fri\_c0\_100\_5      & 100   & 5 & synthetic\\
641\_fri\_c1\_500\_10     & 500   & 10 & synthetic\\
647\_fri\_c1\_250\_10     & 250   & 10 & synthetic\\
654\_fri\_c0\_500\_10     & 500   & 10 & synthetic\\
665\_sleuth\_case2002     & 147   & 6 & real-world\\
666\_rmftsa\_ladata       & 508   & 10 & real-world\\
\bottomrule
    \end{tabular}
\end{table}

\newpage
\section{Results and Discussions}\label{sec:results}
We compare lexicase-based selection methods and two traditional selection methods combined with different down-sampling strategies on a wide range of symbolic regression problems under various conditions. In detail, we compare tournament selection (denoted as \textit{tourn}), fitness-proportionate selection (\textit{fps}), $\epsilon$-lexicase selection (\textit{$\epsilon$-lex}), $\epsilon$-plexicase selection (\textit{$\epsilon$-plex}), batch-tournament selection (\textit{batch-tourn}), and batch-$\epsilon$-lexicase selection (\textit{batch-$\epsilon$-lex}) in combination with no down-sampling (\textit{nds}), with random down-sampling (\textit{rds}) and with informed down-sampling (\textit{ids}).

For the sake of simplicity, only the results for the best batch size found are shown for batch-tournament selection and batch-$\epsilon$-lexicase selection, meaning the batch size with the lowest average MSE on the validation cases for each problem.
 
First, we show and discuss the results for a given evaluation budget. Then, we present the results for a given running time. A statistical analysis of the results is given in Appendix~\ref{appendix:sig_test}.

\subsection{Results for runs with a given maximum number of evaluations}

Usually, selection methods are compared over a fixed evaluation budget because fitness evaluations are considered the most expensive operation~\cite{boldi2024informed}. This means for runs without down-sampling that each selection method searches for $100$ generations to find a solution. For runs with random down-sampling ($d = 0.1$) the search is performed for $1,000$ generations, and for runs with informed down-sampling ($d = 0.1$, $s = 0.01$, $g = 10$) for $991$ generations (see Sect.~\ref{sec:experimental_setting}). 

\begin{figure}[b]
    \centering
    \includegraphics[scale = 0.68]{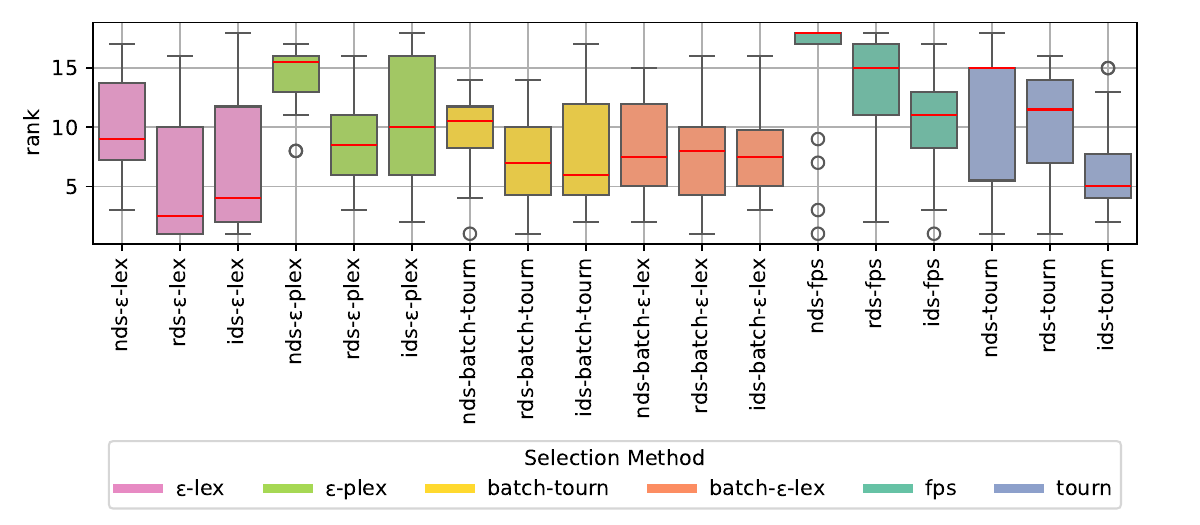}
    \caption{\textbf{Performance} ranking of the selection methods \textbf{for a given evaluation budget}. The performance on each problem is measured in terms of the median MSE on the test cases (smaller is better).}
    \label{fig:rank_eval_budget}
\end{figure}

Figure~\ref{fig:rank_eval_budget} plots the performance rankings of the selection methods for all 26 problems following the approach of \citeN{Orzechowski.07022018} for comparing multiple algorithms over many datasets. 
For each method, we measured the relative performance on a problem by comparing the median MSE on the test cases over all 30 runs.
The down-sampling strategy is indicated by a prefix (\textit{nds} for no down-sampling, \textit{rds} for random down-sampling and \textit{ids} for informed down-sampling).

We observe that almost all methods perform better in combination with a down-sampling strategy. The overall best result is achieved by rds-$\epsilon$-lex and the second best by ids-$\epsilon$-lex. Interestingly, we observe that ids-tourn and ids-fps perform significantly better than their counterpart without 
down-sampling, with ids-tourn being the third best method in terms of the median ranking.

\begin{landscape}
\begin{table} 
    \centering
    \vspace*{25mm}
    \caption{Median MSE on the test cases for each selection method on all 26 problems. The results refer to the best solution found for \textbf{a given evaluation budget}. Best results are shown in bold font. The results are rounded to two decimal places.} \label{tab:results_eval_budget}
 \begin{adjustbox}{width=1.44\textwidth}
\begin{tabular}{l|rrr|rrr|rrr|rrr|rrr|rrr}
\toprule
& \multicolumn{3}{c}{$\epsilon$-lex} & \multicolumn{3}{c}{$\epsilon$-plex} & \multicolumn{3}{c}{batch-tourn} & \multicolumn{3}{c}{batch-$\epsilon$-lex} & \multicolumn{3}{c}{fps} & \multicolumn{3}{c}{tourn} \\
problem & nds & rds & ids & nds & rds & ids & nds & rds & ids & nds & rds & ids & nds & rds & ids & nds & rds & ids \\
\midrule

R-1027 & 0.53 & 0.51 & 0.38 & 0.51 & 0.46 & 0.40 & 0.40 & \textbf{0.33} & 0.37 & 0.35 & 0.36 & 0.36 & 0.66 & 0.57 & 0.54 & 0.57 & 0.39 & 0.38 \\
R-1028 & 0.55 & 0.50 & 0.49 & 0.50 & 0.48 & 0.51 & 0.47 & 0.45 & 0.48 & 0.45 & \textbf{0.44} & 0.47 & 0.62 & 0.48 & 0.49 & 0.48 & 0.45 & 0.46 \\
R-1030 & 2.85 & 2.81 & 2.96 & 3.20 & 2.95 & 3.06 & 2.78 & 2.81 & 2.99 & 2.83 & 2.75 & 2.77 & 3.28 & 2.99 & 2.99 & 2.78 & \textbf{2.71} & 2.88 \\
R-207  & 1.19e+07 & 1.05e+07 & 1.36e+07 &  1.15e+07 & 1.25e+07 & 1.05e+07  & 1.14e+07 & 1.14e+07 & 1.06e+07 & 1.22e+07 & \textbf{8.92e+06} & 1.16e+07 & 1.42e+07 & 1.03e+07 & 9.88e+06 & 9.32e+06 & 1.14e+07 & 1.02e+07 \\
R-230 & 4270.43 & \textbf{3567.15} & 4268.94 & 8685.30 & 4436.89 & 4022.64 & 6131.70 & 5600.98 & 5339.38 & 6049.93 & 5527.29 & 4262.77 & 9720.09 & 6145.74 & 5322.44 & 7565.13 & 7669.26 & 4829.54 \\
R-4544 & 0.48 & \textbf{0.34} & 0.35 & 0.52 & 0.39 & \textbf{0.34} & 0.44 & 0.35 & 0.39 & 0.43 & 0.46 & 0.49 & 0.56 & 0.51 & 0.43 & 0.57 & 0.45 & 0.43 \\
R-505 & 46.92 & 18.78 & 13.60 & 75.75 & 31.28 & 19.66 & 37.20 & \textbf{12.07} & 13.27 & 27.08 & 13.76 & 14.99 & 112.80 & 24.99 & 19.39 & 38.20 & 17.96 & 13.37 \\
R-519 & 2.90 & 3.10 & 3.00 & 2.90 & 2.95 & 3.31 & 2.84 & 2.84 & 3.17 & 2.74 & 2.71 & 2.99 & \textbf{2.49} & 2.50 & 2.63 & 2.50 & 2.55 & 2.84 \\
R-522 & 0.78 & 0.75 & 0.78 & 0.77 & 0.78 & 0.84 & \textbf{0.69} & 0.72 & 0.78 & 0.72 & 0.71 & 0.73 & 0.75 & 0.76 & 0.77 & 0.71 & 0.71 & 0.72 \\
R-523 & 1.12 & 1.11 & 1.25 & 1.12 & 1.05 & 1.11 & 1.05 & \textbf{1.01} & 1.15 & 1.06 & 1.15 & 1.07 & 1.07 & \textbf{1.01} & 1.08 & 1.03 & 1.11 & 1.12 \\
R-560 & 30.70 & 24.13 & 15.03 & 42.09 & 23.82 & 28.09 & 22.54 & 8.21 & 16.96 & 22.22 & \textbf{7.30} & 10.31 & 50.87 & 28.45 & 22.58 & 27.36 & 20.37 & 18.11 \\
S-581 & 0.13 & 0.09 & \textbf{0.08} & 0.47 & 0.12 & 0.15 & 0.21 & 0.19 & 0.13 & 0.18 & 0.23 & 0.15 & 0.76 & 0.75 & 0.27 & 0.58 & 0.44 & 0.11 \\
S-589 & 0.20 & \textbf{0.07} & \textbf{0.07} & 0.31 & 0.18 & 0.16 & 0.24 & 0.20 & 0.15 & 0.24 & 0.21 & 0.20 & 0.44 & 0.33 & 0.24 & 0.29 & 0.23 & 0.17 \\
S-591 & 0.32 & \textbf{0.26} & 0.41 & 0.43 & 0.45 & 0.63 & 0.42 & 0.34 & 0.39 & 0.32 & 0.36 & 0.41 & 0.46 & 0.44 & 0.36 & 0.42 & 0.44 & 0.30 \\
S-606 & 0.19 & 0.08 & \textbf{0.07} & 0.33 & 0.17 & 0.14 & 0.22 & 0.22 & 0.12 & 0.23 & 0.22 & 0.14 & 0.40 & 0.37 & 0.26 & 0.30 & 0.32 & 0.11 \\
S-607 & 0.13 & \textbf{0.07} & 0.08 & 0.51 & 0.11 & 0.12 & 0.21 & 0.20 & 0.12 & 0.25 & 0.14 & 0.13 & 0.84 & 0.79 & 0.70 & 0.56 & 0.32 & 0.13 \\
S-615 & 0.19 & \textbf{0.14} & 0.15 & 0.63 & 0.15 & 0.22 & 0.21 & 0.19 & 0.17 & 0.17 & 0.20 & 0.19 & 0.75 & 0.76 & 0.19 & 0.54 & 0.38 & 0.16 \\
S-617 & 0.16 & 0.10 & \textbf{0.08} & 0.38 & 0.15 & 0.12 & 0.18 & 0.18 & 0.11 & 0.18 & 0.17 & 0.12 & 0.58 & 0.37 & 0.15 & 0.33 & 0.24 & 0.09 \\
S-621 & 0.39 & 0.52 & 0.54 & 0.51 & 0.53 & 0.60 & 0.57 & 0.50 & 0.63 & 0.36 & 0.46 & 0.57 & 0.64 & 0.56 & \textbf{0.35} & 0.51 & 0.59 & 0.42 \\
S-623 & 0.15 & \textbf{0.06} & 0.07 & 0.36 & 0.12 & 0.11 & 0.15 & 0.13 & 0.08 & 0.15 & 0.15 & 0.10 & 0.70 & 0.53 & 0.21 & 0.43 & 0.20 & 0.09 \\
S-624 & 0.44 & \textbf{0.28} & 0.48 & 0.50 & 0.52 & 0.66 & 0.41 & 0.43 & 0.53 & 0.35 & 0.42 & 0.53 & 0.53 & 0.49 & 0.46 & 0.42 & 0.50 & 0.40 \\
S-641 & 0.16 & \textbf{0.07} & \textbf{0.07} & 0.35 & 0.17 & 0.16 & 0.23 & 0.21 & 0.16 & 0.22 & 0.16 & 0.18 & 0.65 & 0.34 & 0.28 & 0.29 & 0.27 & 0.16 \\
S-647 & 0.21 & \textbf{0.13} & 0.16 & 0.44 & 0.17 & 0.21 & 0.22 & 0.28 & 0.21 & 0.19 & 0.25 & 0.18 & 0.74 & 0.48 & 0.26 & 0.35 & 0.29 & 0.18 \\
S-654 & 0.27 & \textbf{0.09} & \textbf{0.09} & 0.36 & 0.25 & 0.16 & 0.27 & 0.18 & 0.10 & 0.23 & 0.26 & 0.23 & 0.58 & 0.42 & 0.30 & 0.34 & 0.32 & 0.13 \\
R-665 & 81.60 & 81.58 & 96.96 & 79.03 & 75.41 & 101.67 & 76.31 & 94.13 & 94.13 & 70.86 & 79.19 & 83.57 & 67.47 & 75.71 & 99.07 & \textbf{61.02} & 62.76 & 92.08 \\
R-666 & 7.10 & 6.60 & 6.36 & 7.57 & 6.52 & 6.85 & 6.52 & \textbf{5.01} & 6.16 & 7.20 & 5.89 & 5.94 & 13.22 & 6.84 & 6.14 & 10.18 & 5.94 & 7.27 \\

\bottomrule
\end{tabular}
\end{adjustbox}
\end{table}
\end{landscape}

Since studying the ranking of each method only provides a summarized overview of their performance, we also show their actual performance in terms of the median MSE on the test cases for each problem in Table~\ref{tab:results_eval_budget}. The problems are abbreviated by their number (see Table~\ref{tab:problems} for the full names) and the prefix \textit{R} is added for real-world problems and \textit{S} for synthetic problems. Boxplots of the distribution of the results for each problem are in Appendix~\ref{appendix:eval_budget}.

According to the median MSE on the test cases, rds-$\epsilon$-lex is the best performing method on 11 out of 20 problems and ids-$\epsilon$-lex performs best on 6 out of 20 problems (on three problems, rds-$\epsilon$-lex and ids-$\epsilon$-lex perform equally well). Interestingly, on 12 out of 13 synthetic problems either rds-$\epsilon$-lex or ids-$\epsilon$-lex performed best. Of the 13 real-world problems, rds-batch-tourn performed best on 4, rds-batch-$\epsilon$-lex on 3, and rds-$\epsilon$-lex only on 2.
Overall, variants using random down-sampling performed best on 19 out of 26 problems.

\begin{figure}[h]
    \centering
    \includegraphics[scale = 0.68]{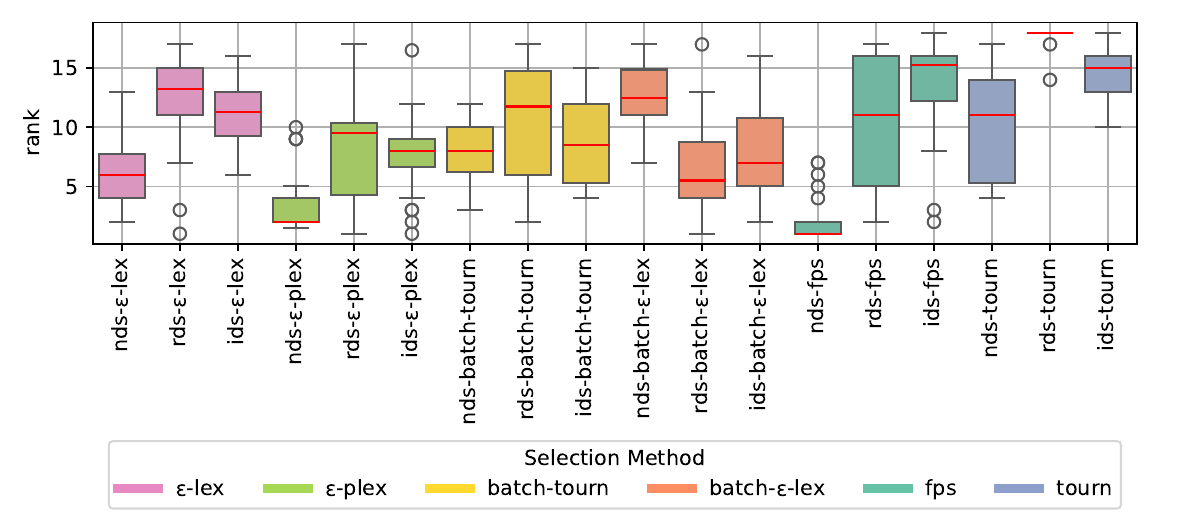}
    \caption{Ranking of the median \textbf{tree size} for all selection methods \textbf{for a given evaluation budget}. The tree size is measured in terms of the median tree size of the final population (smaller is better).}
    \label{fig:tree_size_eval_budget}
\end{figure}

\citeN{la2023flexible} argue that for symbolic regression tasks one should not only take into account the performance of a model but also its simplicity, which is typically approximated by the size of a model. Therefore, we measured the median tree size of the final population for each run. Then, we compared the median population tree size of each method to derive their ranking for each problem (smaller median tree size is better), which is shown in Figure~\ref{fig:tree_size_eval_budget}.

Tournament selection generates populations with a large median tree size, especially rds-tourn. Fitness-proportionate selection generates the smallest individuals for the variant without down-sampling, but significantly larger ones when combined with random or informed down-sampling. 
Apart from that, we found that the trees generated by the best performing methods rds-$\epsilon$-lex and ids-$\epsilon$-lex are relatively large while the ones generated by the worst performing methods nds-fps, and nds-$\epsilon$-plex are small. We observed that rds-batch-$\epsilon$-lex, ids-batch-$\epsilon$-lex, and ids-batch-tourn generate populations with relatively small individuals while still performing relatively well. 

Furthermore, we observed that the tree size usually increases over time for all selection methods~\cite{geiger2024comprehensive}. However, evaluating the performance of individuals is more expensive if their tree size is large. As trees grow over time, generations get more expensive. This is not taken into account when comparing methods only for a given evaluation budget.

\begin{figure}[h]
    \centering
    \includegraphics[scale = 0.68]{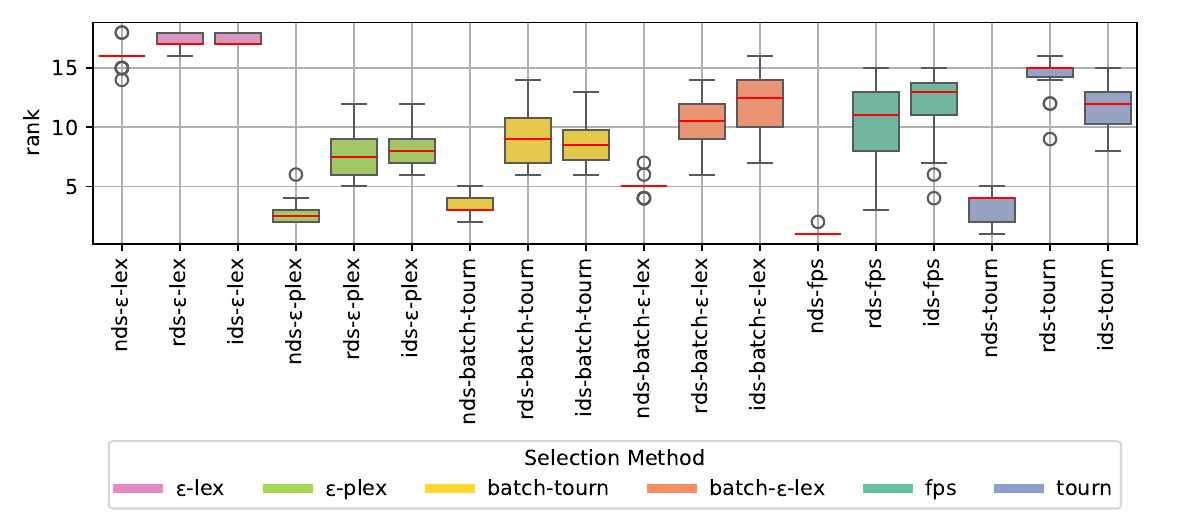}
    \caption{Ranking of the median \textbf{run time} of each selection method \textbf{for a given evaluation budget} (less is better).}
    \label{fig:rank_duration_eval_budget}
\end{figure}

Additionally, the run time of the selection methods varies greatly. In Figure~\ref{fig:rank_duration_eval_budget}, we can see the ranking of the median time each method required to perform the runs for each problem (less time is better). 

For all methods, the variants without down-sampling are faster. We can see that the fastest method is nds-fps, followed by nds-$\epsilon$-plex, nds-batch-tourn and nds-tourn. The slowest methods are the $\epsilon$-lex variants. This means that some methods could perform more generations in the same time than others. Therefore, we are not only comparing the selection methods over a given evaluation budget but also over a given time period. 

To sum up, we found that $\epsilon$-lexicase selection in combination with random and informed down-sampling performs best for a given evaluation budget, especially for synthetic problems. For real-world problems, batch-tournament and batch-$\epsilon$-lexicase selection combined with random down-sampling performed well. Further, we found that the combination with down-sampling improved the performance for almost all selection methods. 
We observed that the median tree size of the generated populations as well as the run time varies greatly between methods. Therefore, we are not only comparing the methods for a given maximum number of evaluations but also for a given amount of time.

\subsection{Results for runs with a given maximum running time}

Now, we compare all methods for a given running time instead of a given maximum number of evaluations. This means that some methods run for more generations than others. We analyze the results for different run times to observe if there are differences in the relative performance of each method over time. We have chosen time budgets of 24h, 1h, and 15 minutes based on our observations in our previous paper~\cite{geiger2024comprehensive}. We observed that most changes in the performance of the methods occurred within the first hour. The relative performance remained similar at the end of the observed 24h period.

\begin{figure}[h]
    \centering
    \includegraphics[scale = 0.69]{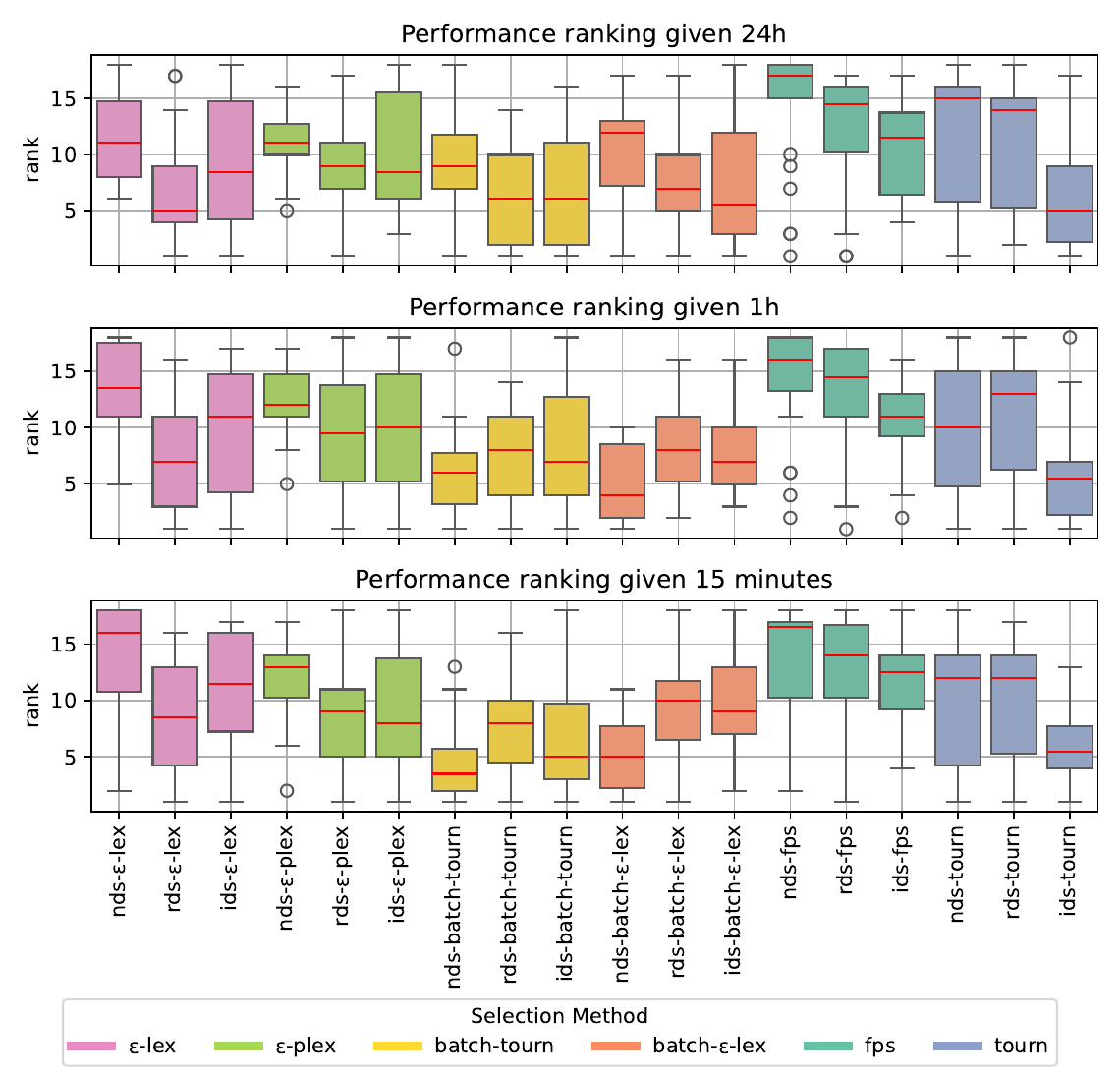}
    \caption{\textbf{Performance} ranking of the selection methods given 24h, 1h or 15 minutes. The performance on each problem is measured in terms of the median MSE on the test cases (smaller is better).}
    \label{fig:rank_time}
\end{figure}

In Figure~\ref{fig:rank_time}, the performance ranking of each method is shown after a running time of 24h, 1h or 15 minutes. 
First, we compare the best solution found within a running time of 24h. For all methods, the combination with a down-sampling method improved the performance. For 4 out of 6 methods, the combination with informed down-sampling performed best. Overall, the best median ranking within 24h was achieved by ids-tourn and rds-$\epsilon$-lex, followed by ids-batch-$\epsilon$-lex. The worst performing methods are nds-fps and nds-tourn.

For practitioners, the time to find a solution for a problem is often strictly limited. Therefore, we also analyze the results found after a shorter running time. 
The rankings of the results found within 1h differ only slightly from the ones for 24h. One major difference is that the relative performance of batch-tourn and batch-$\epsilon$-lex without down-sampling is much better now. According to the median ranking after 1h, the best performing method is nds-batch-$\epsilon$-lex, followed by ids-tourn, and nds-batch-tourn.

We observed further changes in the performance rankings when we limited the time even more. Looking at the results after 15 minutes, we can see that nds-batch-tourn is the best performing method now, followed by ids-batch-tourn and nds-batch-$\epsilon$-lex.

To get a more detailed view at the results, Table~\ref{tab:results_24h} shows the median MSE of all methods on all problems after 24h. Although ids-tourn and rds-$\epsilon$-lex performed best according to the median ranking, they were only the best performing methods on 6 and 3 problems respectively. However, ids-batch-tourn performed best on 7 problems, and rds-batch-tourn on 6 problems. All in all, variants using informed down-sampling achieved the best results on 12 out of 13 synthetic problems and variants using random down-sampling achieved the best results on 10 out of 13 real-world problems. The detailed results for the time budgets of 1h and 15 minutes are shown in Appendix~\ref{appendix:time_period}.

\begin{figure}[h]
    \centering
    \includegraphics[scale=0.68]{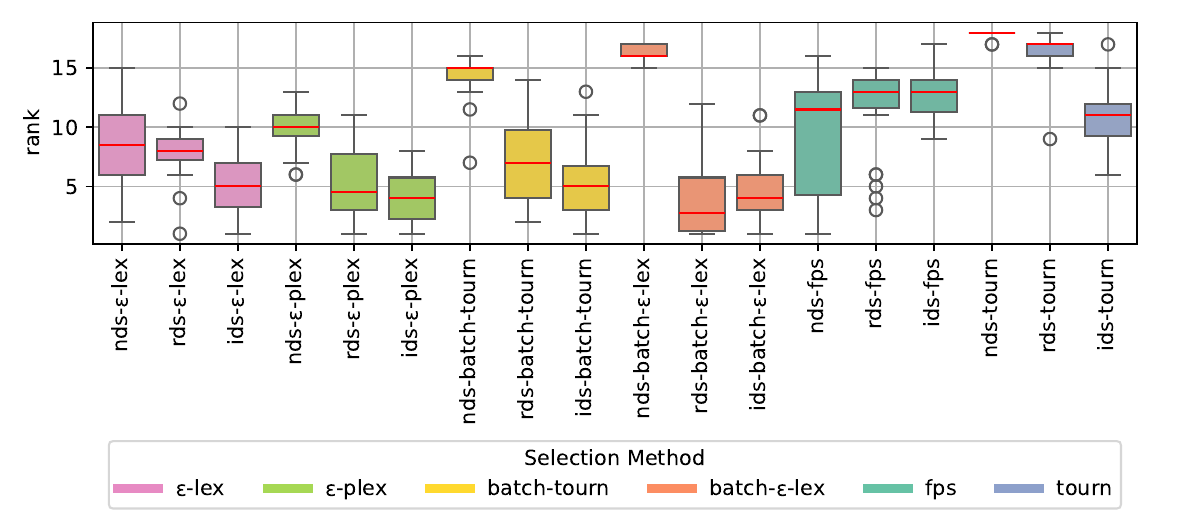}
    \caption{Ranking of the median \textbf{tree size} for all selection methods \textbf{given 24h}. The tree size is measured in terms of the median tree size of the final population (smaller is better).}
    \label{fig:rank_size_24h}
\end{figure}

As before, we are not only analyzing the performance but also at the tree sizes generated by each method. Figure~\ref{fig:rank_size_24h} shows the ranking of the median population tree size for the population after 24h. The populations with the smallest individuals are generated by rds-batch-$\epsilon$-lex, followed by ids-batch-$\epsilon$-lex, and ids-$\epsilon$-plex. 
Ids-batch-$\epsilon$-lex was the third best performing method (according to Fig.~\ref{fig:rank_time}) suggesting that this method offers a good compromise between size and performance. 
Interestingly, in most cases, the variants with down-sampling generate populations with smaller individuals than their counterpart without down-sampling. 

Additionally, the rankings of the tree sizes for the populations given 1h or 15 minutes are shown in Appendix~\ref{appendix:tree_size}. Again, tournament selection generates larger trees compared to the other methods. The populations with the smallest median tree size are generated by nds-$\epsilon$-lex. 

Further, the tables in Appendix~\ref{appendix:gen} show the median number of performed generations for a given running time. Obviously, the variants using down-sampling searched for a larger number of generations compared to the methods without down-sampling. Generally, the $\epsilon$-lexicase variants search for the lowest median number of generations. 

\begin{landscape}
\begin{table} 
    \centering
    \vspace*{25mm}
    \caption{Median MSE on the test cases for each selection method on all 26 problems. The results refer to the best solution found after \textbf{24h}. Best results are shown in bold font. The results are rounded to two decimal places.} \label{tab:results_24h}
 \begin{adjustbox}{width=1.44\textwidth}
\begin{tabular}{l|rrr|rrr|rrr|rrr|rrr|rrr}
\toprule
& \multicolumn{3}{c}{$\epsilon$-lex} & \multicolumn{3}{c}{$\epsilon$-plex} & \multicolumn{3}{c}{batch-tourn} & \multicolumn{3}{c}{batch-$\epsilon$-lex}& \multicolumn{3}{c}{fps} & \multicolumn{3}{c}{tourn} \\
problem & nds & rds & ids & nds & rds & ids & nds & rds & ids & nds & rds & ids & nds & rds & ids & nds & rds & ids \\
\midrule
R-1027 & 0.59 & 0.45 & 0.36 & 0.43 & 0.36 & 0.39 & 0.37 & 0.33 & 0.34 & 0.35 & \textbf{0.32} & 0.34 & 0.42 & 0.36 & 0.34 & 0.36 & 0.38 & 0.34 \\
R-1028 & 0.50 & 0.49 & 0.48 & 0.46 & 0.45 & 0.46 & 0.43 & 0.44 & 0.45 & \textbf{0.42} & 0.43 & 0.46 & 0.47 & 0.47 & 0.47 & 0.44 & \textbf{0.42} & 0.45 \\
R-1030 & 2.86 & 2.82 & 2.84 & 2.81 & 2.80 & 2.75 & \textbf{2.61} & 2.64 & 2.76 & 2.67 & 2.73 & 2.87 & 2.99 & 2.86 & 2.83 & 2.65 & 2.71 & 2.80 \\
R-207 & 9.49e+06 & 1.01e+07 & 1.20e+07 & 9.39e+06 & 9.16e+06 & 1.15e+07 & 8.90e+06 &  \textbf{6.83e+06} & 7.51e+06 & 1.02e+07 & 8.35e+06 & 9.54e+06 & 9.33e+06 & 8.45e+06 & 1.05e+07 & 8.14e+06 & 7.17e+06 & 8.22e+06  \\
R-230 & 4297.13 & \textbf{3247.03} & 3419.23 & 4920.95 & 4263.54 & 4705.39 & 8421.04 & 4316.76 & 5056.89 & 4683.25 & 4322.90 & 5403.67 & 4242.00 & 4123.41 & 3543.73 & 5003.36 & 3494.20 & 3733.44 \\
R-4544 & 0.32 & 0.31 & 0.32 & 0.33 & \textbf{0.29} & 0.33 & 0.33 & 0.30 & 0.33 & 0.33 & 0.32 & 0.32 & 0.49 & 0.46 & 0.36 & 0.47 & 0.39 & 0.40 \\
R-505 & 11.02 & 10.17 & 6.37 & 11.90 & 6.31 & 7.41 & 7.98 & \textbf{3.17} & 6.15 & 9.30 & 4.00 & 5.34 & 12.22 & 8.52 & 6.07 & 8.69 & 13.26 & 4.50 \\
R-519 & 3.23 & 2.91 & 3.51 & 2.85 & 3.05 & 3.26 & 2.81 & 2.93 & 3.07 & 2.68 & 3.01 & 3.23 & 2.61 & \textbf{2.55} & 2.85 & 2.58 & 2.80 & 3.03 \\
R-522 & 0.72 & 0.75 & 0.83 & 0.71 & 0.77 & 0.86 & 0.68 & 0.68 & 0.79 & \textbf{0.67} & 0.71 & 0.77 & 0.71 & 0.72 & 0.70 & \textbf{0.67} & 0.69 & 0.74 \\
R-523 & 1.10 & 1.06 & 1.12 & 1.10 & 1.09 & 1.09 & 1.08 & 1.07 & 1.10 & 1.11 & 1.16 & 1.17 & \textbf{1.01} & 1.05 & 1.13 & 1.06 & 1.05 & 1.12 \\
R-560 & 11.68 & 2.45 & 3.44 & 9.59 & 2.06 & 2.62 & 4.64 & \textbf{1.92} & 2.28 & 13.74 & 2.14 & 2.87 & 18.46 & 13.67 & 11.01 & 9.66 & 6.88 & 2.74 \\
S-581 & 0.08 & 0.06 & 0.06 & 0.09 & 0.10 & 0.07 & 0.07 & 0.08 & \textbf{0.05} & 0.11 & 0.08 & \textbf{0.05} & 0.55 & 0.18 & 0.09 & 0.30 & 0.31 & \textbf{0.05} \\
S-589 & 0.11 & 0.05 & \textbf{0.04} & 0.11 & 0.08 & 0.05 & 0.15 & 0.14 & \textbf{0.04} & 0.14 & 0.12 & 0.12 & 0.32 & 0.31 & 0.21 & 0.21 & 0.20 & 0.13 \\
S-591 & 0.25 & 0.17 & 0.30 & 0.27 & 0.31 & 0.43 & 0.23 & 0.19 & 0.31 & 0.24 & 0.20 & 0.20 & 0.38 & 0.34 & 0.22 & 0.39 & 0.36 & \textbf{0.14} \\
S-606 & 0.11 & 0.06 & 0.06 & 0.10 & 0.08 & 0.07 & 0.14 & 0.18 & \textbf{0.04} & 0.17 & 0.10 & 0.05 & 0.33 & 0.31 & 0.18 & 0.23 & 0.26 & 0.05 \\
S-607 & 0.08 & 0.06 & \textbf{0.05} & 0.09 & 0.08 & 0.06 & 0.08 & 0.10 & \textbf{0.05} & 0.09 & 0.08 & \textbf{0.05} & 0.75 & 0.31 & 0.08 & 0.26 & 0.18 & 0.06 \\
S-615 & 0.11 & 0.09 & \textbf{0.07} & 0.11 & 0.08 & \textbf{0.07} & 0.10 & 0.10 & 0.08 & 0.11 & 0.09 & \textbf{0.07} & 0.62 & 0.34 & 0.10 & 0.45 & 0.30 & 0.08 \\
S-617 & 0.08 & 0.07 & 0.06 & 0.09 & 0.08 & 0.06 & 0.08 & 0.10 & 0.05 & 0.10 & 0.10 & 0.05 & 0.17 & 0.12 & 0.09 & 0.24 & 0.19 & \textbf{0.04} \\
S-621 & 0.33 & \textbf{0.12} & 0.42 & 0.38 & 0.57 & 0.63 & 0.34 & \textbf{0.12} & 0.39 & 0.36 & 0.17 & 0.24 & 0.56 & 0.47 & 0.16 & 0.56 & 0.53 & \textbf{0.12} \\
S-623 & 0.09 & 0.05 & 0.05 & 0.09 & 0.07 & 0.06 & 0.07 & 0.08 & \textbf{0.04} & 0.07 & 0.08 & \textbf{0.04} & 0.52 & 0.17 & 0.07 & 0.20 & 0.16 & 0.05 \\
S-624 & 0.24 & 0.14 & 0.36 & 0.29 & 0.39 & 0.61 & 0.32 & 0.16 & 0.37 & 0.25 & \textbf{0.13} & 0.26 & 0.40 & 0.37 & 0.15 & 0.40 & 0.38 & 0.14 \\
S-641 & 0.07 & 0.05 & 0.04 & 0.08 & 0.07 & 0.04 & 0.14 & 0.10 & \textbf{0.03} & 0.11 & 0.06 & 0.04 & 0.27 & 0.22 & 0.23 & 0.21 & 0.20 & 0.06 \\
S-647 & 0.08 & 0.09 & 0.06 & 0.12 & 0.06 & 0.06 & 0.13 & 0.11 & 0.07 & 0.17 & 0.08 & 0.06 & 0.25 & 0.24 & 0.12 & 0.25 & 0.21 & \textbf{0.05} \\
S-654 & 0.10 & \textbf{0.07} & 0.08 & 0.12 & 0.09 & 0.09 & 0.09 & \textbf{0.07} & \textbf{0.07} & 0.12 & 0.08 & \textbf{0.07} & 0.35 & 0.27 & 0.11 & 0.27 & 0.25 & \textbf{0.07} \\
R-665 & 93.82 & 76.34 & 116.30 & 78.78 & 87.83 & 108.50 & 103.81 & 82.36 & 89.56 & 86.34 & 91.18 & 92.17 & 71.18 & \textbf{69.17} & 106.33 & 71.15 & 80.85 & 110.90 \\
R-666 & 6.85 & 5.27 & 6.87 & 6.60 & 6.27 & 6.23 & 5.17 & \textbf{4.82} & 5.64 & 6.75 & 5.62 & 4.85 & 6.75 & 6.44 & 5.97 & 6.69 & 5.67 & 5.78 \\

\bottomrule
\end{tabular}
\end{adjustbox}
\end{table}
\end{landscape}

% \begin{figure}[h]
%     \centering
%     \includegraphics[scale = 0.68]{figures/rank_1h.pdf}
%     \caption{\textbf{Performance} ranking of the selection methods \textbf{given 1h}. The performance on each problem is measured in terms of the median MSE on the test cases (smaller is better).}
%     \label{fig:rank_1h}
% \end{figure}

% \begin{figure}[h]
%     \centering
%     \includegraphics[scale = 0.68]{figures/rank_15min.pdf}
%     \caption{\textbf{Performance} ranking of the selection methods \textbf{given 15 minutes}. The performance on each problem is measured in terms of the median MSE on the test cases (smaller is better).}
%     \label{fig:rank_15min}
% \end{figure}

\subsection{Key Findings}
To sum up, we observed that the relative performance of the selection methods as well as the influence of the down-sampling strategies depends on the given termination criteria. For a better overview, we report the median ranking of each method for a given number of evaluation and the studied time budgets in Table~\ref{tab:median_ranking}. 

\begin{table}[h]
    \centering
    \caption{Overview of the median ranking of each method for a given termination criteria. The best median rankings are shown in bold font.}
    \label{tab:median_ranking}
    \begin{tabular}{l|r|r|r|r}
    \toprule
Method & \# Evaluations & 24h & 1h & 15min \\
\midrule
nds-$\epsilon$-lex & 9.0 & 11.0 & 13.5 & 16.0 \\
rds-$\epsilon$-lex & \textbf{2.5} & \textbf{5.0} & 7.0 & 8.5 \\
ids-$\epsilon$-lex & 4.0 & 8.5 & 11.0 & 11.5 \\
nds-$\epsilon$-plex & 15.5 & 11.0 & 12.0 & 13.0 \\
rds-$\epsilon$-plex & 8.5 & 9.0 & 9.5 & 9.0 \\
ids-$\epsilon$-plex & 10.0 & 8.5 & 10.0 & 8.0 \\
nds-batch-tourn & 10.5 & 9.0 & 6.0 & \textbf{3.5} \\
rds-batch-tourn & 7.0 & 6.0 & 8.0 & 8.0 \\
ids-batch-tourn & 6.0 & 6.0 & 7.0 & 5.0 \\
nds-batch-$\epsilon$-lex & 7.5 & 12.0 & \textbf{4.0} & 5.0 \\
rds-batch-$\epsilon$-lex & 8.0 & 7.0 & 8.0 & 10.0 \\
ids-batch-$\epsilon$-lex & 7.5 & 5.5 & 7.0 & 9.0 \\
nds-fps & 18.0 & 17.0 & 16.0 & 16.5 \\
rds-fps & 15.0 & 14.5 & 14.5 & 14.0 \\
ids-fps & 11.0 & 11.5 & 11.0 & 12.5 \\
nds-tourn & 15.0 & 15.0 & 10.0 & 12.0 \\
rds-tourn & 11.5 & 14.0 & 13.0 & 12.0 \\
ids-tourn & 5.0 & \textbf{5.0} & 5.5 & 5.5 \\
\bottomrule
    \end{tabular}
\end{table}

Down-sampling generally improves the performance. Only for shorter running times, the lexicase variants using batches of training cases perform better without down-sampling. It is worth noting that informed down-sampling has the greatest influence on the performance of tournament selection. Tournament selection without down-sampling or with random down-sampling is among the worst performing methods in all settings while tournament selection combined with informed down-sampling performs well in all studied settings and is even the best performing method for a given running time of 24h. We assume that the subset of training cases created with informed down-sampling leads to an average error that weights different behaviors more equally, since training cases that test the same behavior are excluded. 

Further, we observe that $\epsilon$-lexicase selection combined with random down-sampling is the best performing method for a given number of evaluations and a running time of 24h.  However, the median ranking is worse when the $\epsilon$-lexicase variants are compared for a given running time, especially if the running time is short. We assume that this is due to the long running times of $\epsilon$-lexicase (see Fig.~\ref{fig:rank_duration_eval_budget}), which means that the search is performed for a lower number of generations compared to other methods (see Appendix~\ref{appendix:gen}). 

The median ranking of batch-tournament selection and batch-$\epsilon$-lexicase selection gets better for shorter running times and they even outperform the other methods for a given time of 1h or 15 minutes. This indicates that the use of batches of training cases is especially useful if the given time to find a solution is  short. We assume that this is because the batch variants evaluate individuals from the beginning on several training cases, making the selection process less dependent on the order of the training cases~\cite{Aenugu.2019}. 

\section{Conclusions}\label{sec:conclusions}

We compared relevant lexicase variants in combination with different down-sampling strategies on a wide range of symbolic regression problems. 
%das würde ich rauslassen. Der unterschied zu unserer vorherigen Publikation gehört eigentlich nicht in die conclusions, sondern nur in die intro. und da ist es ja schon. We including random as well as informed down-sampling, by extending our problem set, and by providing detailed results for all studied problems. 
As a baseline, we included fitness-proportionate selection and tournament selection. To our knowledge, this is the first work studying the influence of informed down-sampling on various selection methods on a wide range of symbolic regression problems. We not only compared the selection method over a fixed number of evaluations, as usually done in the literature, but also over different running times  as users usually have limited time to solve their problems. Therefore, this work provides users with a comprehensive guide for choosing the best selection method for their individual problem. Furthermore, we analyzed the influence of the selection methods on the solution size.

We found that the relative performance of each selection method depends on the given setting. For a fixed number of evaluations, $\epsilon$-lexicase selection in combination with random or informed down-sampling performed best, especially for synthetic problems. For almost all selection methods, the combination with a down-sampling strategy increased the performance. Furthermore, we observed that the median population tree size as well as the median runtime differs between methods. This highlights the importance of comparing the selection methods also for a given running time and not just for a given evaluation budget, as it is usually the case in literature. 

Therefore, we studied the solutions found after 24h, as well as after 1h and 15 minutes. For a given running time of 24h, the best performing methods are tournament selection in combination with informed down-sampling and $\epsilon$-lexicase selection in combination with random down-sampling. If running time is more limited (1h or less) lexicase variants using batches of training cases outperform all other methods. 

To sum up, choosing the right selection method depends on the given setting. For a given evaluation budget, we recommend using $\epsilon$-lexicase selection in combination with a down-sampling strategy. If there is a need for a quick solution, batch-$\epsilon$-lexicase selection or batch-tournament selection lead to the best solution. Interestingly, the combination of tournament selection with informed down-sampling performed well in all settings. However, the detailed results show that while there are general trends, the performance of each method depends on the problem at hand. 

In future work, we will extend this study by analyzing other population dynamics next to tree size like diversity or specialist selection. Additionally, we will study the influence of informed down-sampling on tournament selection in the domain of symbolic regression further. 

\bibliographystyle{ACM-Reference-Format}
\bibliography{main}

\clearpage
\appendix
\section{Results for a given maximum number of evaluations}\label{appendix:eval_budget}

\begin{figure}[h]
    \centering
    \includegraphics[scale=0.9]{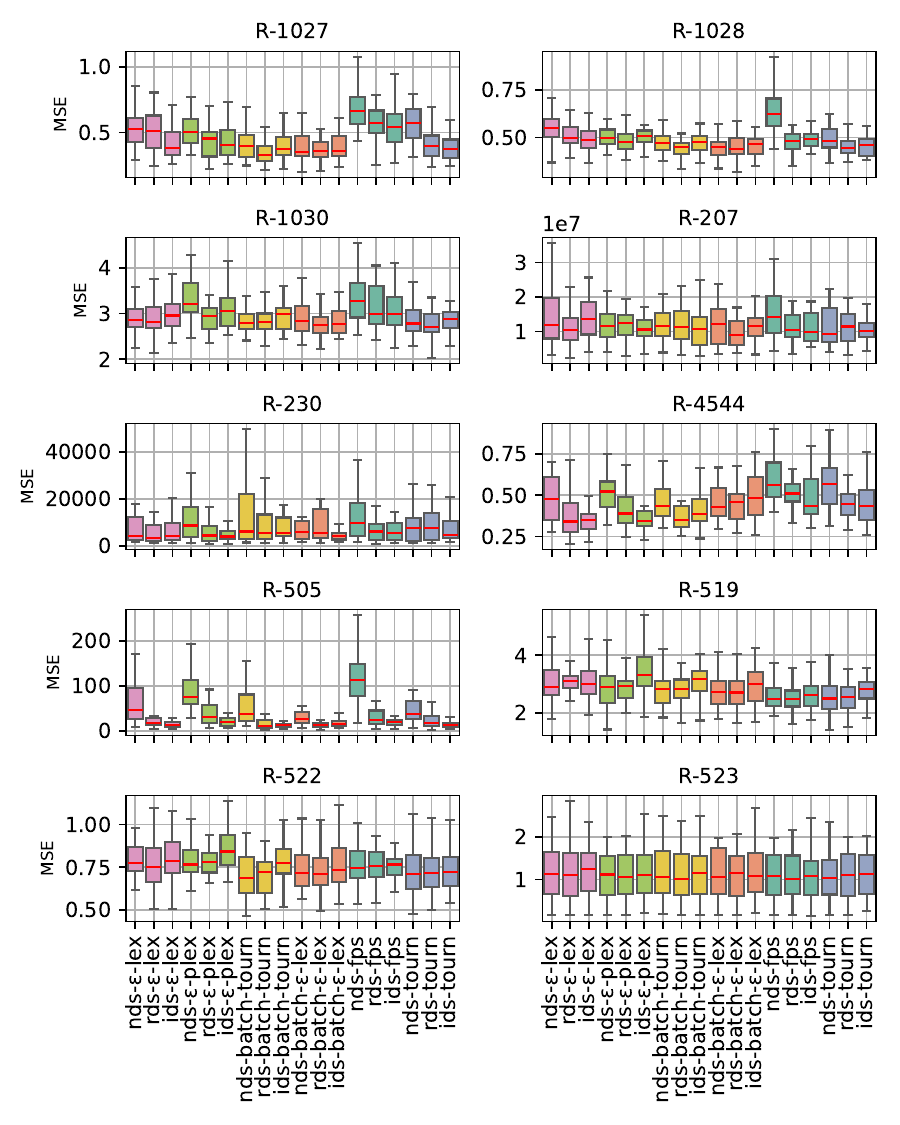}
    \caption{MSE on the test cases for each selection method on 10 problems. The results refer to the best solution found \textbf{after a given evaluation budget}. Outliers are not shown to improve readability.}
    \label{fig:results_eval_budget_0}
\end{figure}

\begin{figure}[h]
    \centering
    \includegraphics[scale=0.9]{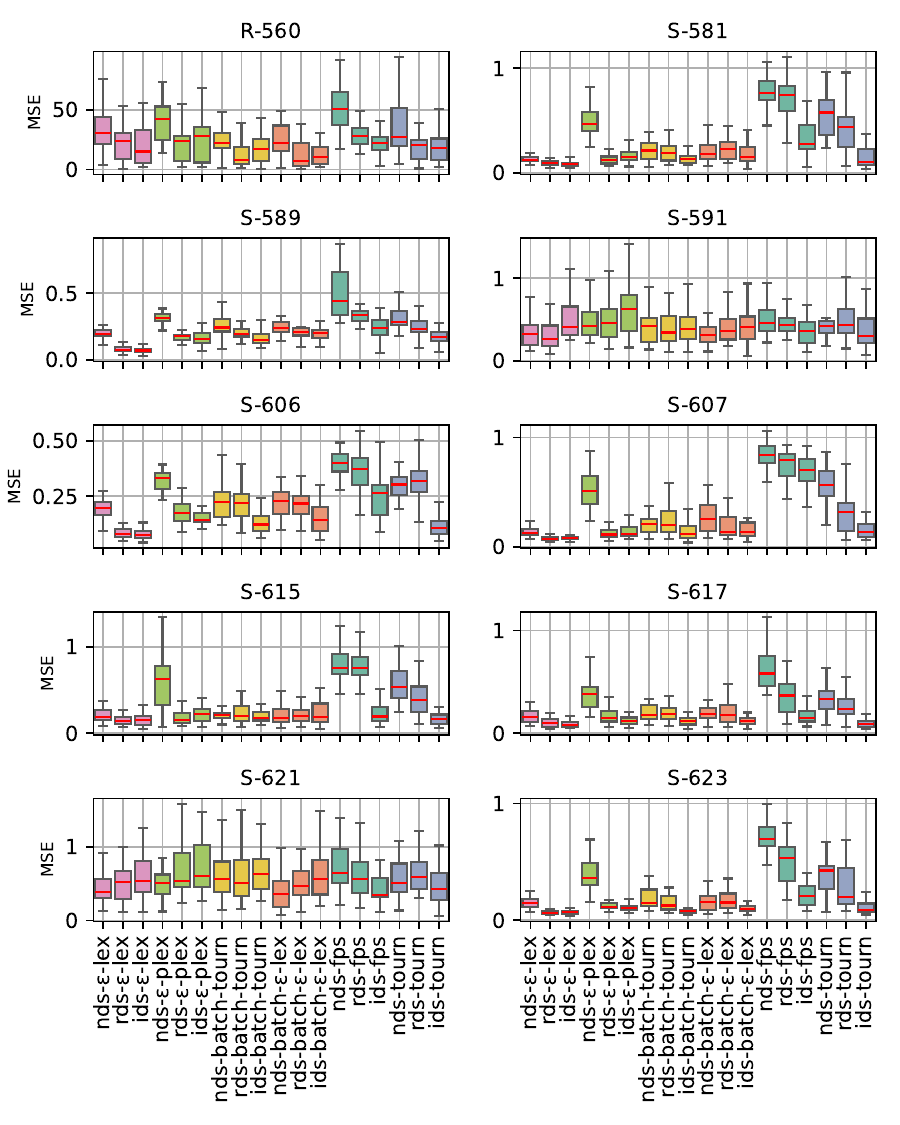}
    \caption{MSE on the test cases for each selection method on 10 problems. The results refer to the best solution found \textbf{after a given evaluation budget}. Outliers are not shown to improve readability.}
    \label{fig:results_eval_budget_8}
\end{figure}

\begin{figure}[h]
    \centering
    \includegraphics[scale=0.9]{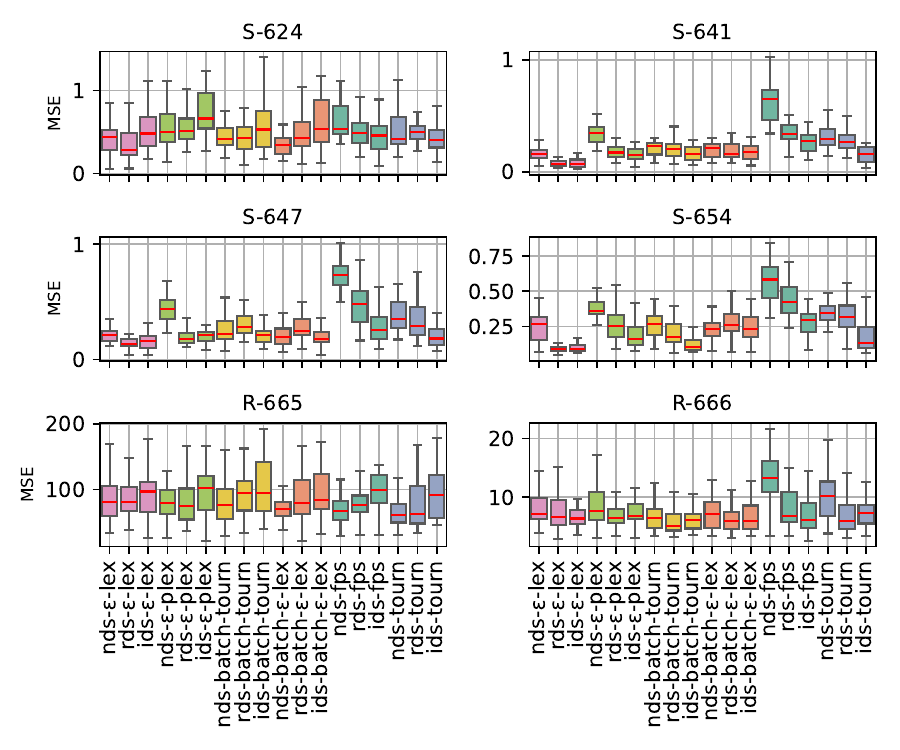}
    \caption{MSE on the test cases for each selection method on 6 problems. The results refer to the best solution found \textbf{after a given evaluation budget}. Outliers are not shown to improve readability.}
    \label{fig:results_eval_budget_16}
\end{figure}

\clearpage
\section{Results for a given run time}\label{appendix:time_period}

\begin{figure}[h]
    \centering
    \includegraphics[scale=0.9]{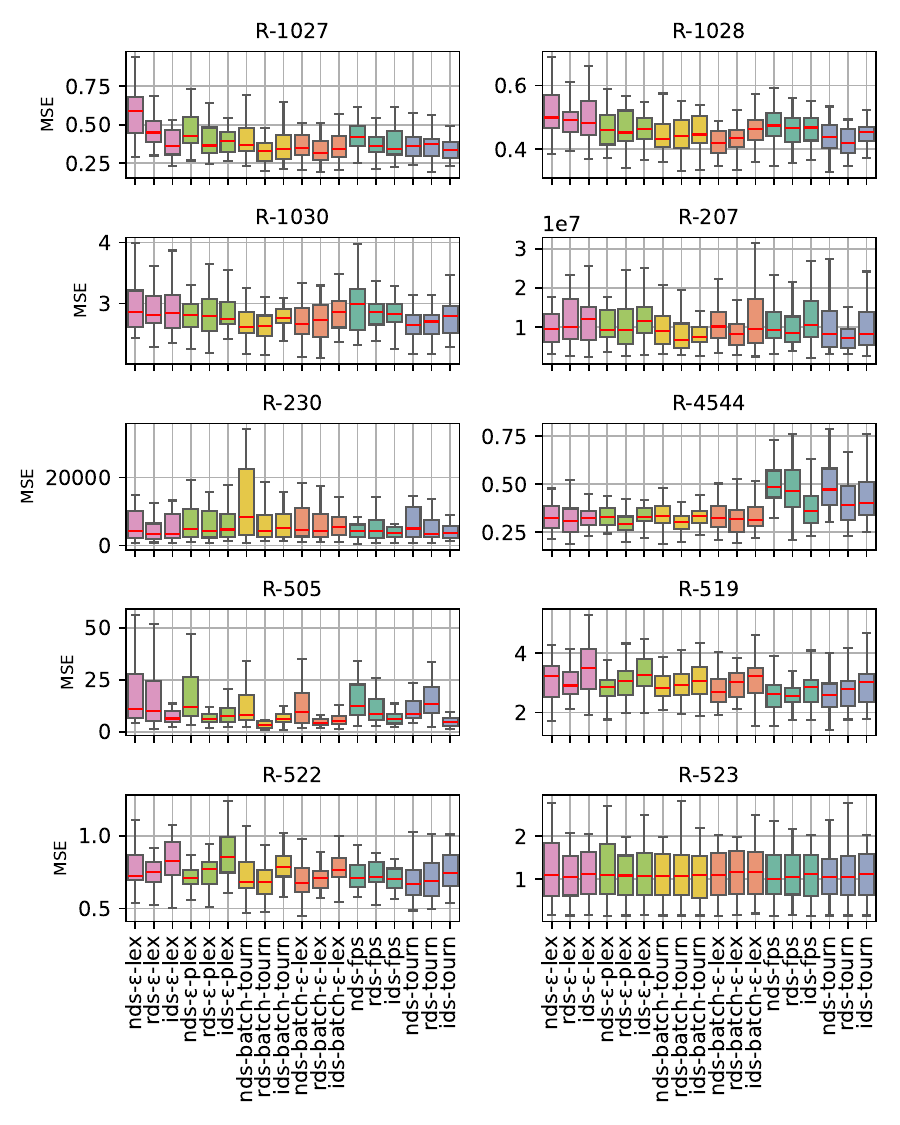}
    \caption{MSE on the test cases for each selection method on 10 problems. The results refer to the best solution found \textbf{after 24h}. Outliers are not shown to improve readability.}
    \label{fig:results_24h_0}
\end{figure}

\begin{figure}[h]
    \centering
    \includegraphics[scale=0.9]{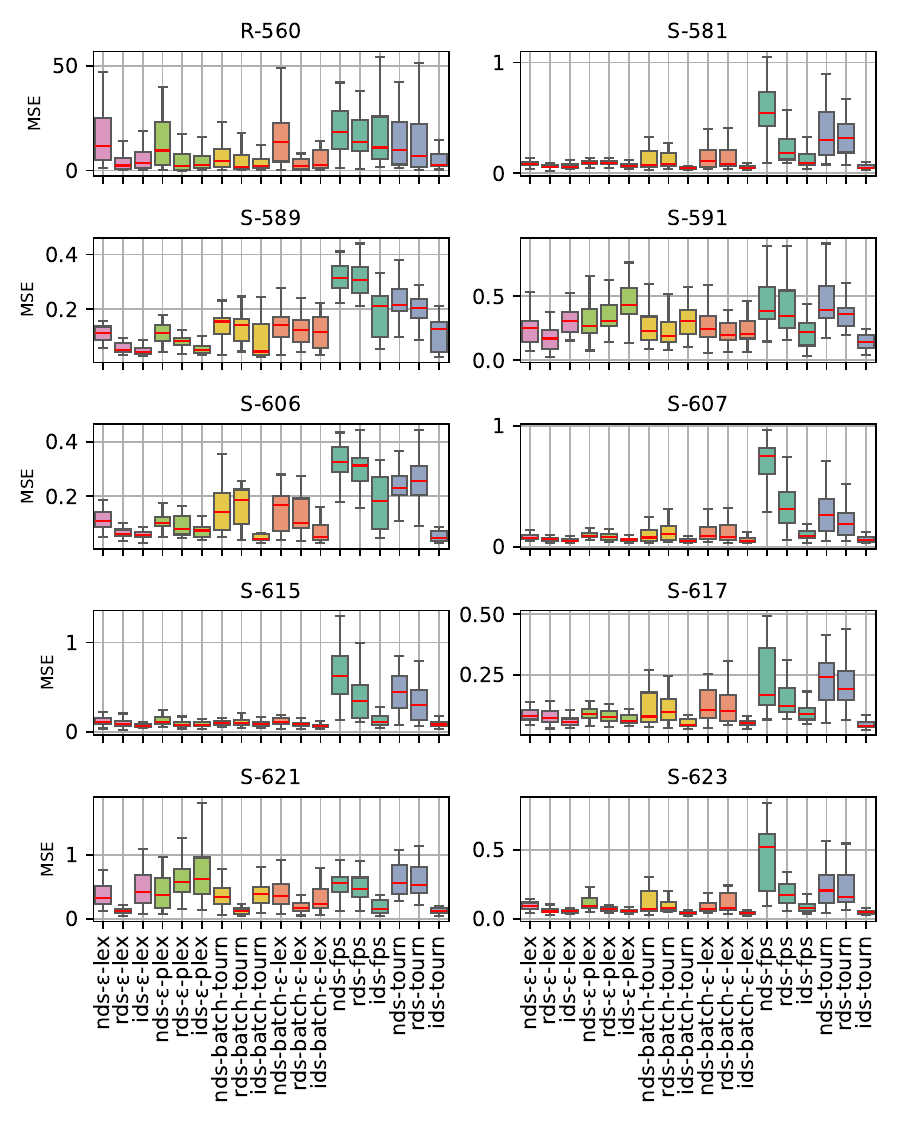}
    \caption{MSE on the test cases for each selection method on 10 problems. The results refer to the best solution found \textbf{after 24h}. Outliers are not shown to improve readability.}
    \label{fig:results_24h_8}
\end{figure}

\begin{figure}[h]
    \centering
    \includegraphics[scale=0.9]{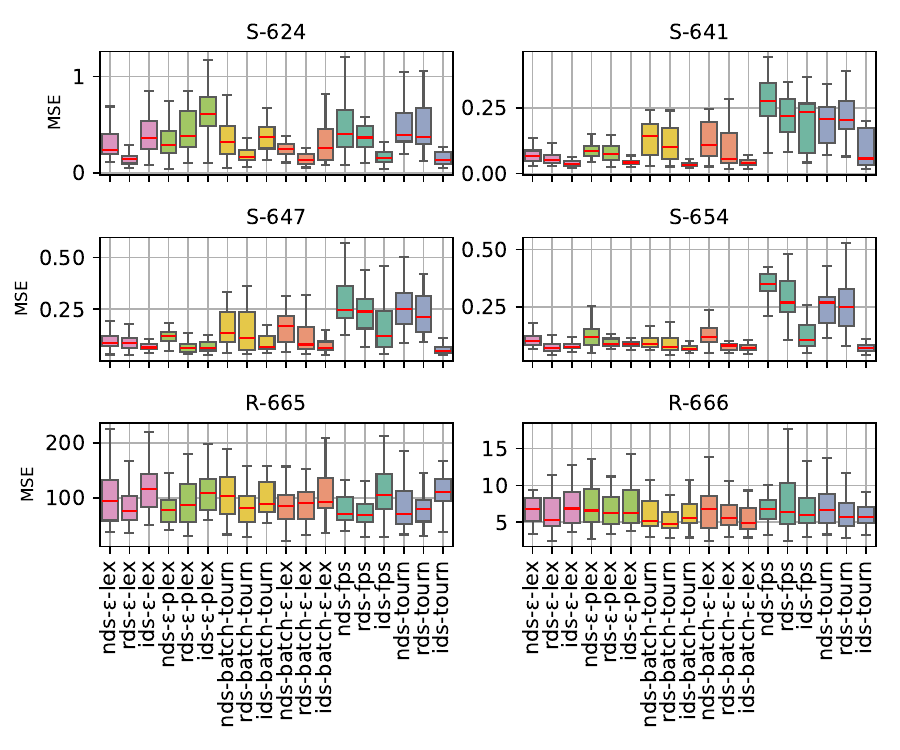}
    \caption{MSE on the test cases for each selection method on 6 problems. The results refer to the best solution found \textbf{after 24h}. Outliers are not shown to improve readability.}
    \label{fig:results_24h_16}
\end{figure}

\begin{figure}[h]
    \centering
    \includegraphics[scale=0.9]{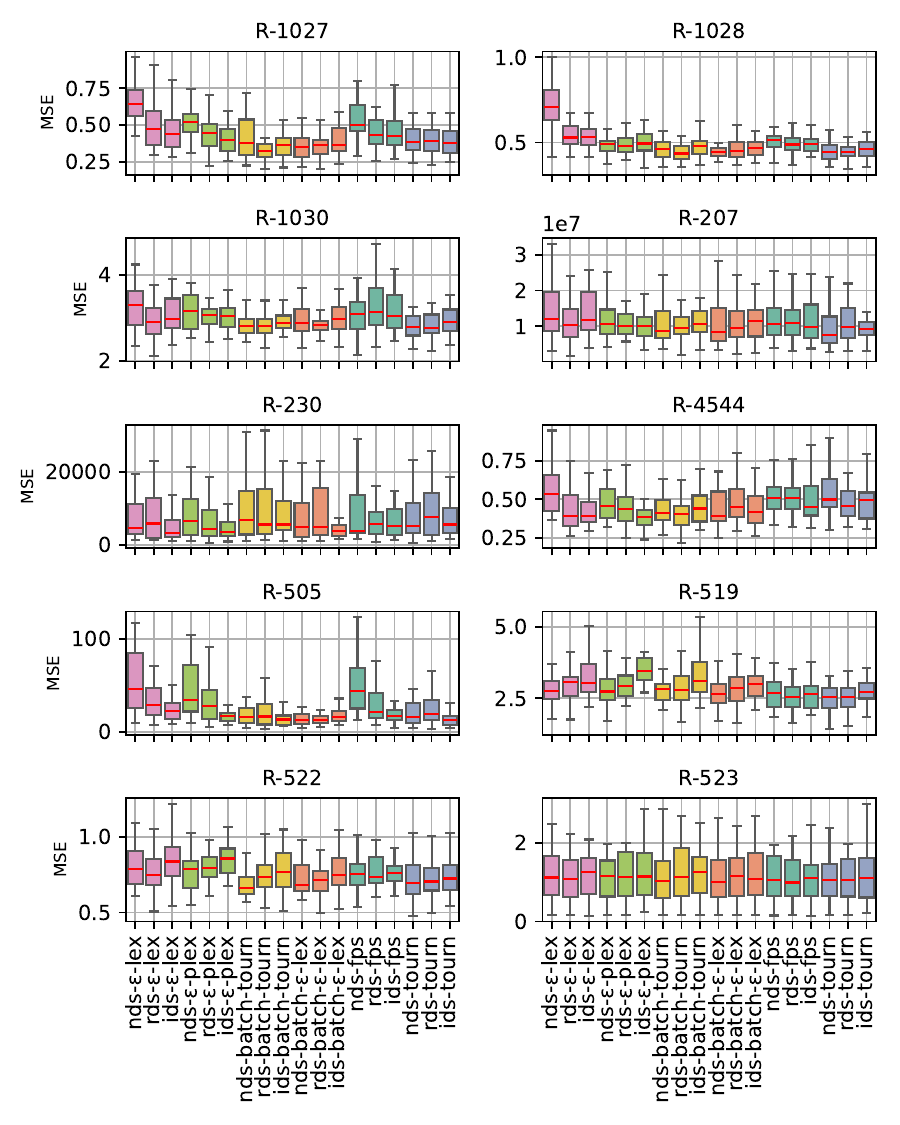}
    \caption{MSE on the test cases for each selection method on 10 problems. The results refer to the best solution found \textbf{after 1h}. Outliers are not shown to improve readability.}
    \label{fig:results_1h_0}
\end{figure}

\begin{figure}[h]
    \centering
    \includegraphics[scale=0.9]{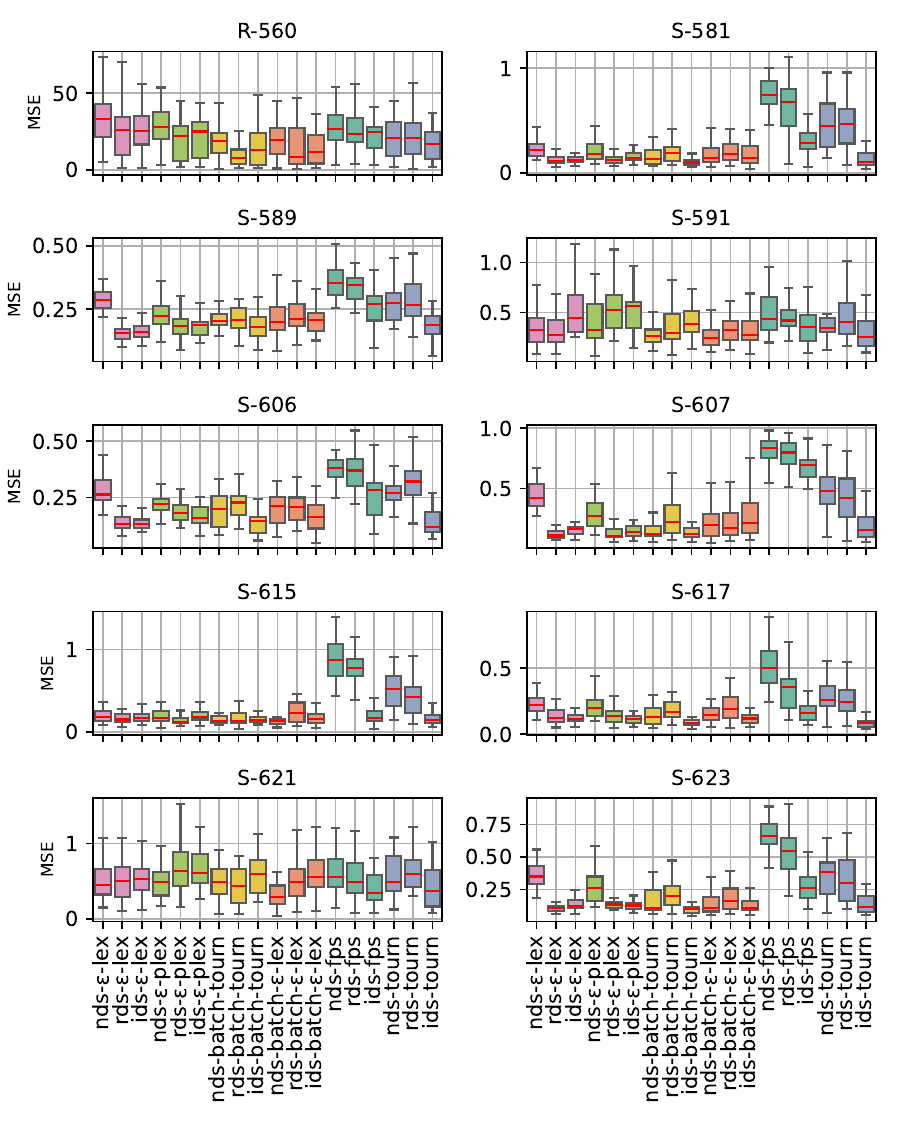}
    \caption{MSE on the test cases for each selection method on 10 problems. The results refer to the best solution found \textbf{after 1h}. Outliers are not shown to improve readability.}
    \label{fig:results_1h_8}
\end{figure}

\begin{figure}[h]
    \centering
    \includegraphics[scale=0.9]{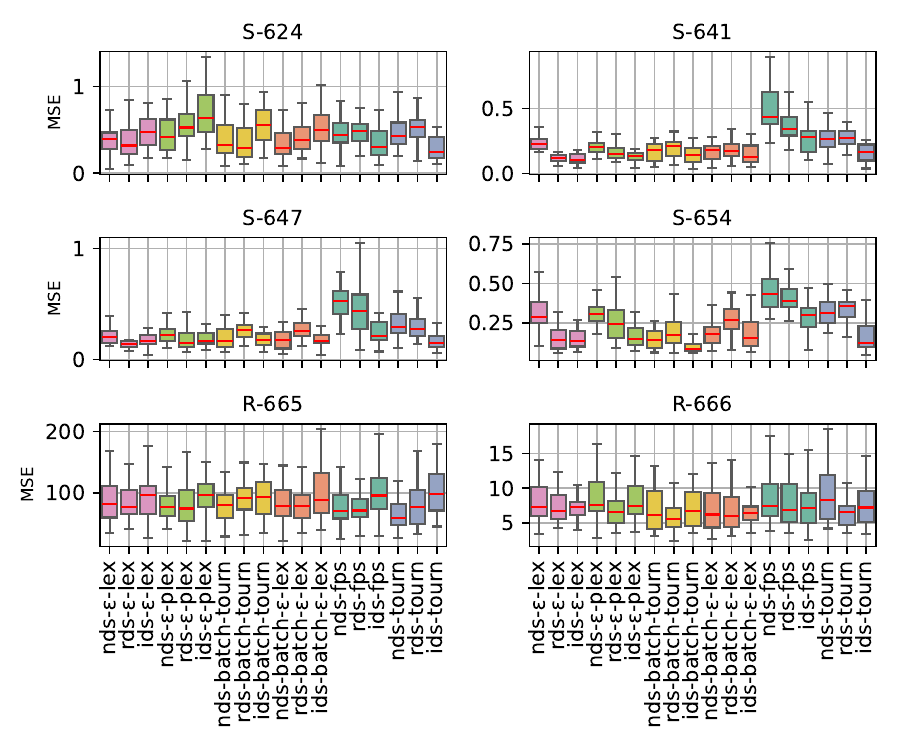}
    \caption{MSE on the test cases for each selection method on 6 problems. The results refer to the best solution found \textbf{after 1h}. Outliers are not shown to improve readability.}
    \label{fig:results_1h_16}
\end{figure}

\begin{landscape}
\begin{table} 
    \centering
    \vspace*{25mm}
    \caption{Median MSE on the test cases for each selection method on all 26 problems. The results refer to the best solution found after \textbf{1h}. Best results are shown in bold font. The results are rounded to two decimal places.} \label{tab:results_1h}
 \begin{adjustbox}{width=1.44\textwidth}
\begin{tabular}{l|rrr|rrr|rrr|rrr|rrr|rrr}
\toprule
 &  \multicolumn{3}{c}{$\epsilon$-lex} & \multicolumn{3}{c}{$\epsilon$-plex} & \multicolumn{3}{c}{batch-tourn} & \multicolumn{3}{c}{batch-$\epsilon$-lex} & \multicolumn{3}{c}{fps} & \multicolumn{3}{c}{tourn} \\
problem & nds & rds & ids & nds & rds & ids & nds & rds & ids & nds & rds & ids & nds & rds & ids & nds & rds & ids \\
\midrule

R-1027 & 0.64 & 0.47 & 0.44 & 0.52 & 0.45 & 0.40 & 0.38 & \textbf{0.32} & 0.36 & 0.35 & 0.36 & 0.36 & 0.50 & 0.43 & 0.43 & 0.38 & 0.39 & 0.38 \\
R-1028 & 0.71 & 0.53 & 0.53 & 0.49 & 0.48 & 0.49 & 0.46 & \textbf{0.43} & 0.48 & 0.45 & 0.45 & 0.47 & 0.52 & 0.49 & 0.49 & 0.45 & 0.45 & 0.46 \\
R-1030 & 3.30 & 2.91 & 2.97 & 3.17 & 3.06 & 3.05 & 2.81 & 2.80 & 2.88 & 2.89 & 2.84 & 2.98 & 3.09 & 3.14 & 3.04 & 2.79 & \textbf{2.76} & 2.91 \\
R-207 & 1.19e+07 & 1.04e+07 & 1.17e+07 & 1.05e+07 & 1.02e+07 & 9.96e+06 & 8.74e+06 & 9.61e+06 & 1.05e+07 & 8.38e+06 & 9.61e+06 & 1.16e+07 & 1.07e+07 & 1.10e+07 & 9.87e+06 & \textbf{7.67e+06} & 9.70e+06 & 9.22e+06 \\
R-230 & 4597.58 & 5877.57 & \textbf{3353.63} & 6552.53 & 4266.53 & 3672.30 & 6863.01 & 5601.98 & 5606.45 & 4890.56 & 5039.28 & 3719.03 & 3825.49 & 5819.78 & 5165.00 & 5182.75 & 7564.11 & 5609.15 \\
R-4544 & 0.53 & \textbf{0.39} & \textbf{0.39} & 0.46 & 0.44 & \textbf{0.39} & 0.41 & 0.40 & 0.44 & \textbf{0.39} & 0.45 & 0.41 & 0.51 & 0.51 & 0.45 & 0.50 & 0.46 & 0.50 \\
R-505 & 45.98 & 28.55 & 22.23 & 34.36 & 27.25 & 17.43 & 15.40 & 16.41 & 13.16 & \textbf{12.46} & 13.06 & 15.60 & 44.20 & 21.08 & 16.95 & 16.03 & 18.86 & 13.12 \\
R-519 & 2.77 & 3.08 & 3.04 & 2.74 & 2.94 & 3.45 & 2.81 & 2.78 & 3.10 & 2.64 & 2.87 & 2.98 & 2.69 & 2.55 & 2.66 & \textbf{2.54} & 2.55 & 2.73 \\
R-522 & 0.79 & 0.75 & 0.84 & 0.79 & 0.80 & 0.86 & \textbf{0.66} & 0.74 & 0.76 & 0.68 & 0.71 & 0.75 & 0.75 & 0.74 & 0.76 & 0.70 & 0.70 & 0.72 \\
R-523 & 1.12 & 1.08 & 1.25 & 1.14 & 1.12 & 1.14 & 1.03 & 1.12 & 1.26 & 1.01 & 1.14 & 1.07 & 1.06 & \textbf{0.99} & 1.11 & 1.05 & 1.06 & 1.10 \\
R-560 & 33.47 & 25.97 & 25.62 & 28.12 & 22.32 & 25.05 & 18.61 & \textbf{7.77} & 13.01 & 19.35 & 8.13 & 11.68 & 26.81 & 23.61 & 24.71 & 21.08 & 20.70 & 16.85 \\
S-581 & 0.22 & 0.12 & 0.12 & 0.18 & 0.12 & 0.14 & 0.14 & 0.19 & \textbf{0.10} & 0.14 & 0.18 & 0.14 & 0.75 & 0.68 & 0.28 & 0.44 & 0.47 & \textbf{0.10} \\
S-589 & 0.29 & \textbf{0.16} & \textbf{0.16} & 0.22 & 0.18 & 0.19 & 0.20 & 0.21 & 0.18 & 0.20 & 0.21 & 0.21 & 0.35 & 0.34 & 0.27 & 0.28 & 0.27 & 0.19 \\
S-591 & 0.32 & 0.27 & 0.44 & 0.33 & 0.53 & 0.57 & 0.26 & 0.30 & 0.38 & \textbf{0.25} & 0.33 & 0.28 & 0.43 & 0.42 & 0.36 & 0.34 & 0.40 & \textbf{0.25} \\
S-606 & 0.26 & 0.13 & 0.13 & 0.22 & 0.18 & 0.16 & 0.20 & 0.23 & 0.14 & 0.21 & 0.21 & 0.16 & 0.38 & 0.37 & 0.28 & 0.27 & 0.32 & \textbf{0.12} \\
S-607 & 0.42 & \textbf{0.11} & 0.16 & 0.27 & \textbf{0.11} & 0.14 & 0.12 & 0.23 & 0.12 & 0.20 & 0.17 & 0.21 & 0.83 & 0.80 & 0.69 & 0.48 & 0.42 & 0.15 \\
S-615 & 0.18 & 0.15 & 0.17 & 0.17 & \textbf{0.12} & 0.18 & 0.13 & 0.14 & 0.15 & 0.13 & 0.23 & 0.16 & 0.87 & 0.78 & 0.16 & 0.52 & 0.42 & 0.14 \\
S-617 & 0.22 & 0.13 & 0.12 & 0.20 & 0.14 & 0.11 & 0.13 & 0.17 & \textbf{0.08} & 0.15 & 0.19 & 0.12 & 0.50 & 0.36 & 0.16 & 0.26 & 0.24 & 0.09 \\
S-621 & 0.44 & 0.50 & 0.53 & 0.49 & 0.63 & 0.61 & 0.49 & 0.44 & 0.60 & \textbf{0.29} & 0.48 & 0.56 & 0.55 & 0.49 & 0.34 & 0.49 & 0.59 & 0.37 \\
S-623 & 0.35 & 0.11 & 0.12 & 0.26 & 0.13 & 0.13 & 0.11 & 0.20 & \textbf{0.10} & \textbf{0.10} & 0.16 & 0.11 & 0.66 & 0.54 & 0.26 & 0.38 & 0.30 & 0.12 \\
S-624 & 0.39 & 0.32 & 0.47 & 0.41 & 0.53 & 0.63 & 0.33 & 0.29 & 0.55 & 0.29 & 0.38 & 0.49 & 0.44 & 0.49 & 0.30 & 0.43 & 0.53 & \textbf{0.24} \\
S-641 & 0.23 & 0.12 & \textbf{0.10} & 0.20 & 0.15 & 0.13 & 0.18 & 0.21 & 0.15 & 0.18 & 0.17 & 0.13 & 0.44 & 0.34 & 0.28 & 0.27 & 0.27 & 0.17 \\
S-647 & 0.20 & \textbf{0.14} & 0.17 & 0.22 & 0.15 & 0.17 & 0.17 & 0.27 & 0.17 & 0.18 & 0.26 & 0.17 & 0.53 & 0.44 & 0.21 & 0.29 & 0.28 & 0.15 \\
S-654 & 0.29 & 0.14 & 0.14 & 0.31 & 0.24 & 0.15 & 0.14 & 0.17 & \textbf{0.09} & 0.18 & 0.27 & 0.16 & 0.43 & 0.39 & 0.30 & 0.31 & 0.36 & 0.12 \\
R-665 & 81.60 & 77.82 & 97.16 & 77.05 & 74.69 & 96.61 & 79.84 & 92.21 & 93.41 & 79.19 & 78.70 & 87.95 & 70.99 & 71.52 & 95.93 & \textbf{58.63} & 77.76 & 98.21 \\
R-666 & 7.32 & 6.71 & 7.24 & 7.51 & 6.52 & 7.38 & 6.07 & \textbf{5.56} & 6.64 & 6.18 & 5.97 & 6.47 & 7.48 & 6.80 & 7.12 & 8.23 & 6.52 & 7.20 \\

\bottomrule
\end{tabular}
\end{adjustbox}
\end{table}
\end{landscape}

\begin{figure}[h]
    \centering
    \includegraphics[scale=0.9]{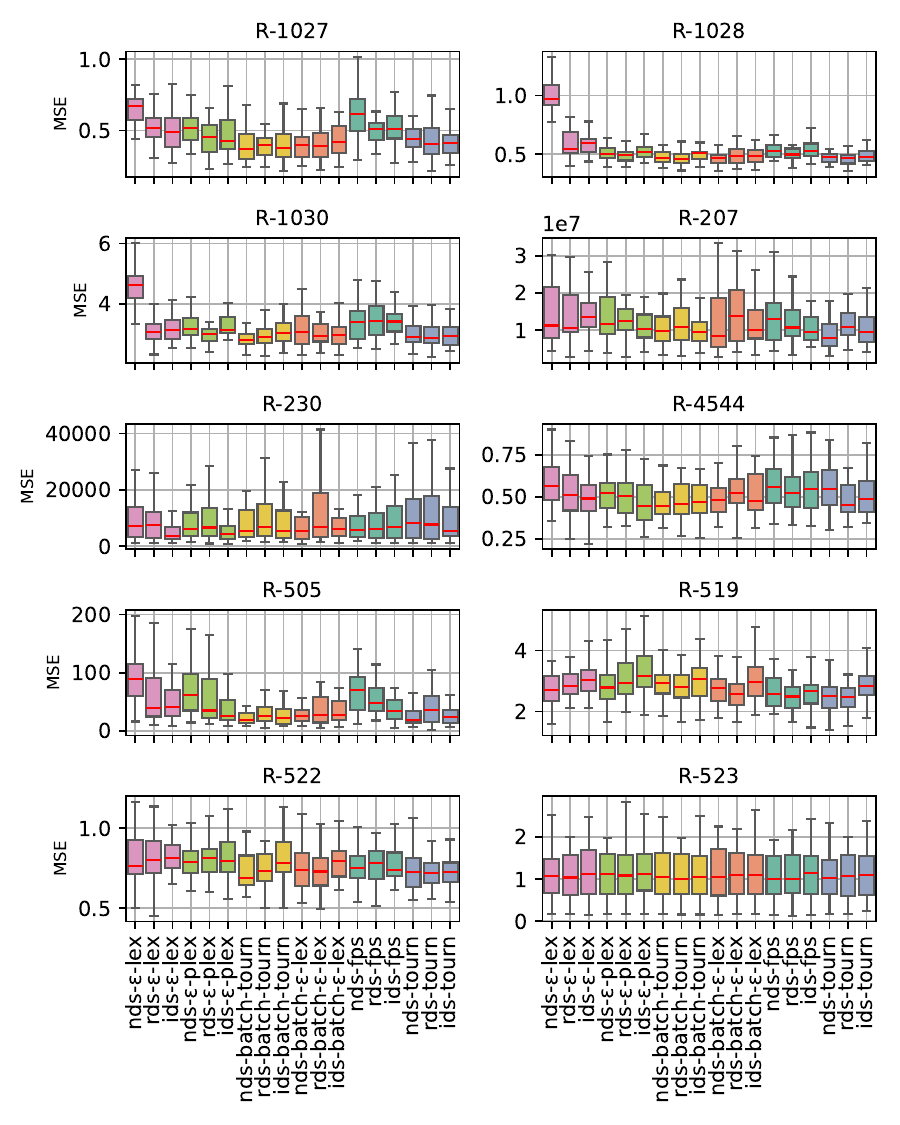}
    \caption{MSE on the test cases for each selection method on 10 problems. The results refer to the best solution found \textbf{after 15 minutes}. Outliers are not shown to improve readability.}
    \label{fig:results_15min_0}
\end{figure}

\begin{figure}[h]
    \centering
    \includegraphics[scale=0.9]{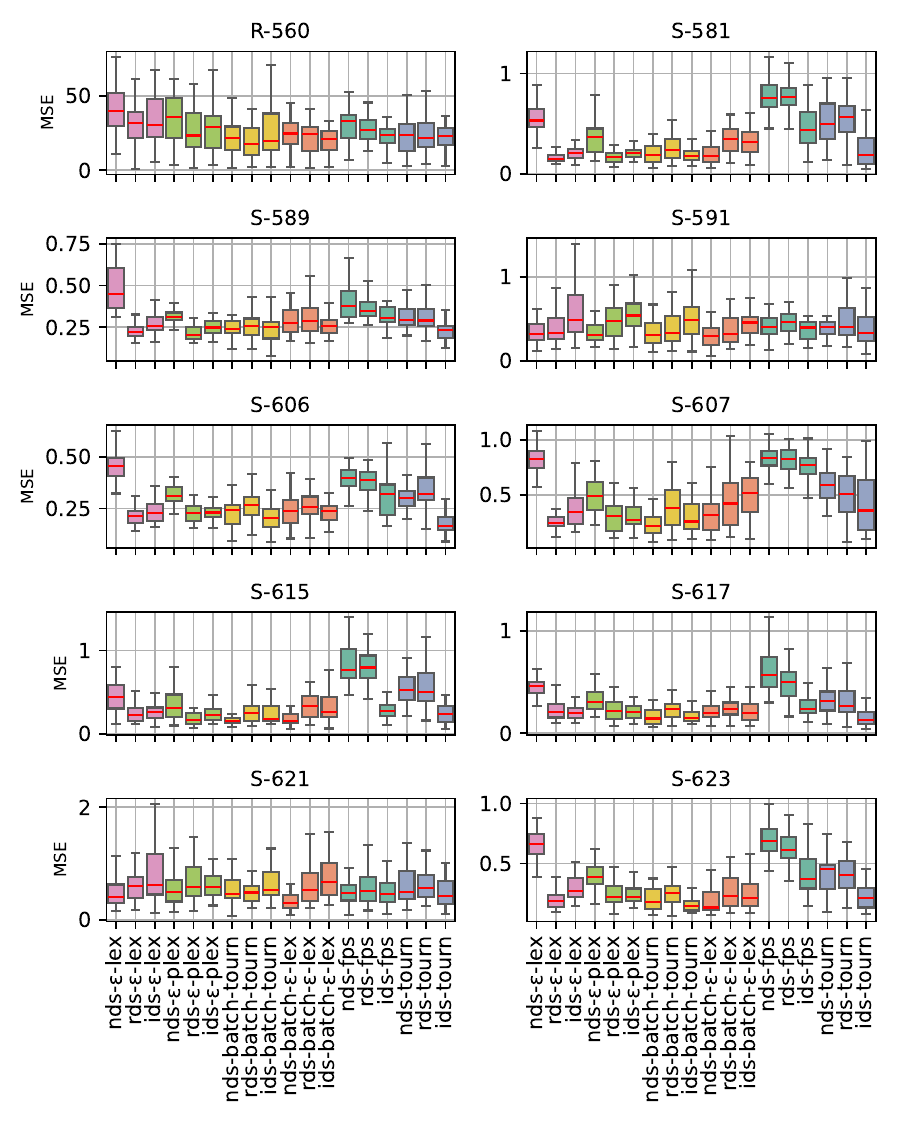}
    \caption{MSE on the test cases for each selection method on 10 problems. The results refer to the best solution found \textbf{after 15 minutes}. Outliers are not shown to improve readability.}
    \label{fig:results_15min_8}
\end{figure}

\begin{figure}[h]
    \centering
    \includegraphics[scale=0.9]{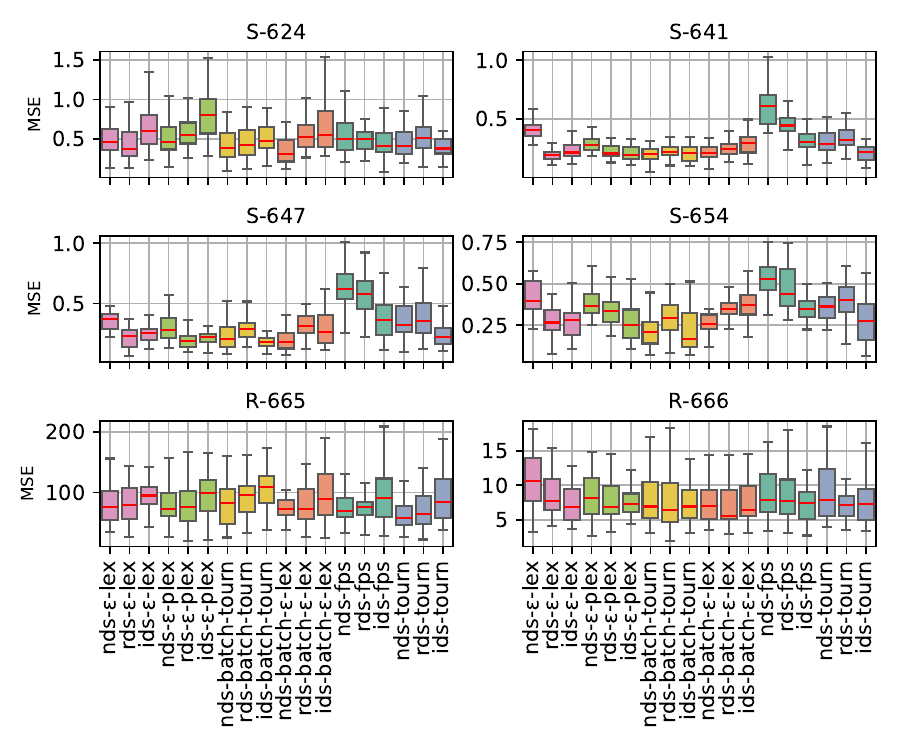}
    \caption{MSE on the test cases for each selection method on 6 problems. The results refer to the best solution found \textbf{after 15 minutes}. Outliers are not shown to improve readability.}
    \label{fig:results_15min_16}
\end{figure}

\begin{landscape}
\begin{table} 
    \centering
    \vspace*{25mm}
    \caption{Median MSE on the test cases for each selection method on all 26 problems. The results refer to the best solution found after \textbf{15 minutes}. Best results are shown in bold font. The results are rounded to two decimal places.} \label{tab:results_15min}
 \begin{adjustbox}{width=1.44\textwidth}
\begin{tabular}{l|rrr|rrr|rrr|rrr|rrr|rrr}
\toprule
 &  \multicolumn{3}{c}{$\epsilon$-lex} & \multicolumn{3}{c}{$\epsilon$-plex} & \multicolumn{3}{c}{batch-tourn} & \multicolumn{3}{c}{batch-$\epsilon$-lex} & \multicolumn{3}{c}{fps} & \multicolumn{3}{c}{tourn} \\
problem & nds & rds & ids & nds & rds & ids & nds & rds & ids & nds & rds & ids & nds & rds & ids & nds & rds & ids \\
\midrule
R-1027 & 0.67 & 0.52 & 0.49 & 0.52 & 0.46 & 0.42 & \textbf{0.37} & 0.40 & 0.38 & 0.40 & 0.39 & 0.42 & 0.61 & 0.51 & 0.51 & 0.44 & 0.41 & 0.41 \\
R-1028 & 0.97 & 0.55 & 0.59 & 0.50 & 0.49 & 0.52 & 0.47 & \textbf{0.46} & 0.51 & 0.47 & 0.48 & 0.48 & 0.53 & 0.50 & 0.53 & 0.47 & \textbf{0.46} & 0.48 \\
R-1030 & 4.62 & 3.07 & 3.12 & 3.17 & 2.99 & 3.13 & \textbf{2.80} & 2.92 & 3.03 & 3.06 & 2.95 & 2.98 & 3.40 & 3.45 & 3.42 & 2.90 & 2.88 & 2.94 \\
R-207 & 1.13e+07 & 1.07e+07 & 1.35e+07 & 1.17e+07  & 1.25e+07 & 1.03e+07 & 9.87e+06& 1.09e+07 & 9.50e+06 & 8.43e+06 &  1.37e+07 & 1.00e+07 & 1.31e+07& 1.07e+07 & 9.50e+06 & \textbf{7.83e+06} &1.09e+07 & 9.57e+06\\
R-230 & 7348.06 & 7693.86 & \textbf{3677.54} & 5991.03 & 6677.00 & 4293.60 & 5371.53 & 6711.46 & 5551.55 & 5437.96 & 6827.64 & 5997.39 & 5704.11 & 6152.14 & 6728.12 & 8405.18 & 7721.95 & 5411.92 \\
R-4544 & 0.56 & 0.51 & 0.49 & 0.52 & 0.50 & \textbf{0.44} & \textbf{0.44} & 0.46 & 0.47 & 0.48 & 0.52 & 0.48 & 0.56 & 0.52 & 0.55 & 0.55 & 0.45 & 0.49 \\
R-505 & 89.09 & 40.13 & 41.29 & 62.03 & 35.27 & 25.84 & 19.31 & 26.11 & 22.88 & 26.11 & 27.29 & 27.80 & 70.65 & 47.93 & 33.83 & \textbf{19.20} & 36.03 & 23.28 \\
R-519 & 2.72 & 2.85 & 3.04 & 2.79 & 2.93 & 3.16 & 2.92 & 2.81 & 3.06 & 2.76 & 2.57 & 2.98 & 2.58 & \textbf{2.49} & 2.67 & 2.52 & \textbf{2.49} & 2.83 \\
R-522 & 0.76 & 0.80 & 0.81 & 0.79 & 0.82 & 0.79 & \textbf{0.69} & 0.73 & 0.78 & 0.74 & 0.73 & 0.80 & 0.75 & 0.78 & 0.74 & 0.73 & 0.72 & 0.72 \\
R-523 & 1.08 & 1.04 & 1.12 & 1.12 & 1.09 & 1.11 & 1.04 & 1.01 & 1.06 & 1.05 & 1.09 & 1.10 & 1.00 & \textbf{0.99} & 1.16 & 1.02 & 1.06 & 1.10 \\
R-560 & 39.99 & 31.94 & 30.22 & 35.55 & 23.31 & 29.30 & 21.75 & \textbf{17.89} & 19.86 & 24.66 & 24.46 & 21.09 & 32.75 & 27.23 & 23.80 & 23.56 & 21.87 & 23.01 \\
S-581 & 0.53 & \textbf{0.14} & 0.21 & 0.36 & 0.17 & 0.21 & 0.19 & 0.23 & 0.17 & 0.18 & 0.35 & 0.32 & 0.76 & 0.77 & 0.44 & 0.50 & 0.57 & 0.18 \\
S-589 & 0.45 & 0.22 & 0.26 & 0.31 & \textbf{0.21} & 0.25 & 0.24 & 0.26 & 0.25 & 0.27 & 0.29 & 0.26 & 0.38 & 0.35 & 0.30 & 0.30 & 0.29 & 0.23 \\
S-591 & 0.32 & 0.33 & 0.49 & 0.31 & 0.48 & 0.54 & 0.31 & 0.33 & 0.49 & \textbf{0.29} & 0.32 & 0.46 & 0.40 & 0.46 & 0.40 & 0.41 & 0.40 & 0.33 \\
S-606 & 0.46 & 0.21 & 0.23 & 0.31 & 0.23 & 0.23 & 0.24 & 0.27 & 0.20 & 0.24 & 0.26 & 0.24 & 0.40 & 0.39 & 0.32 & 0.30 & 0.32 & \textbf{0.16} \\
S-607 & 0.83 & 0.24 & 0.34 & 0.49 & 0.31 & 0.27 & \textbf{0.22} & 0.38 & 0.26 & 0.31 & 0.42 & 0.52 & 0.84 & 0.82 & 0.77 & 0.59 & 0.50 & 0.36 \\
S-615 & 0.45 & 0.23 & 0.26 & 0.31 & 0.16 & 0.23 & \textbf{0.15} & 0.25 & 0.18 & \textbf{0.15} & 0.33 & 0.26 & 0.76 & 0.79 & 0.28 & 0.53 & 0.50 & 0.24 \\
S-617 & 0.46 & 0.21 & 0.19 & 0.30 & 0.22 & 0.21 & 0.14 & 0.24 & 0.14 & 0.19 & 0.24 & 0.20 & 0.56 & 0.50 & 0.23 & 0.31 & 0.26 & \textbf{0.12} \\
S-621 & 0.41 & 0.60 & 0.62 & 0.49 & 0.59 & 0.58 & 0.46 & 0.49 & 0.54 & \textbf{0.31} & 0.53 & 0.68 & 0.48 & 0.52 & 0.45 & 0.50 & 0.57 & 0.42 \\
S-623 & 0.66 & 0.19 & 0.27 & 0.39 & 0.22 & 0.22 & 0.18 & 0.25 & 0.14 & \textbf{0.13} & 0.23 & 0.21 & 0.69 & 0.61 & 0.37 & 0.46 & 0.40 & 0.21 \\
S-624 & 0.46 & 0.37 & 0.60 & 0.46 & 0.55 & 0.81 & 0.39 & 0.42 & 0.47 & \textbf{0.31} & 0.53 & 0.55 & 0.49 & 0.50 & 0.41 & 0.41 & 0.51 & 0.38 \\
S-641 & 0.40 & \textbf{0.19} & 0.21 & 0.28 & 0.20 & \textbf{0.19} & 0.20 & 0.22 & 0.21 & 0.21 & 0.25 & 0.29 & 0.61 & 0.44 & 0.31 & 0.29 & 0.32 & 0.22 \\
S-647 & 0.37 & 0.24 & 0.26 & 0.28 & 0.19 & 0.23 & 0.20 & 0.29 & \textbf{0.18} & \textbf{0.18} & 0.31 & 0.26 & 0.62 & 0.58 & 0.37 & 0.32 & 0.36 & 0.23 \\
S-654 & 0.40 & 0.27 & 0.28 & 0.36 & 0.33 & 0.25 & 0.21 & 0.30 & \textbf{0.17} & 0.26 & 0.35 & 0.37 & 0.53 & 0.44 & 0.35 & 0.36 & 0.40 & 0.27 \\
R-665 & 76.18 & 79.66 & 95.18 & 73.40 & 75.41 & 99.01 & 83.15 & 95.31 & 108.45 & 72.53 & 72.67 & 89.64 & 69.16 & 75.97 & 90.82 & \textbf{57.45} & 64.67 & 83.87 \\
R-666 & 10.59 & 7.73 & 6.86 & 8.21 & 6.90 & 7.28 & 6.92 & 6.46 & 6.93 & 7.03 & \textbf{5.55} & 6.38 & 7.87 & 7.77 & 7.48 & 7.86 & 7.12 & 7.29 \\
\bottomrule
\end{tabular}
\end{adjustbox}
\end{table}
\end{landscape}

\clearpage
\section{Tree sizes for a given running time of 1h or 15 minutes}\label{appendix:tree_size}
\begin{figure}[h]
    \centering
    \includegraphics[scale=0.68]{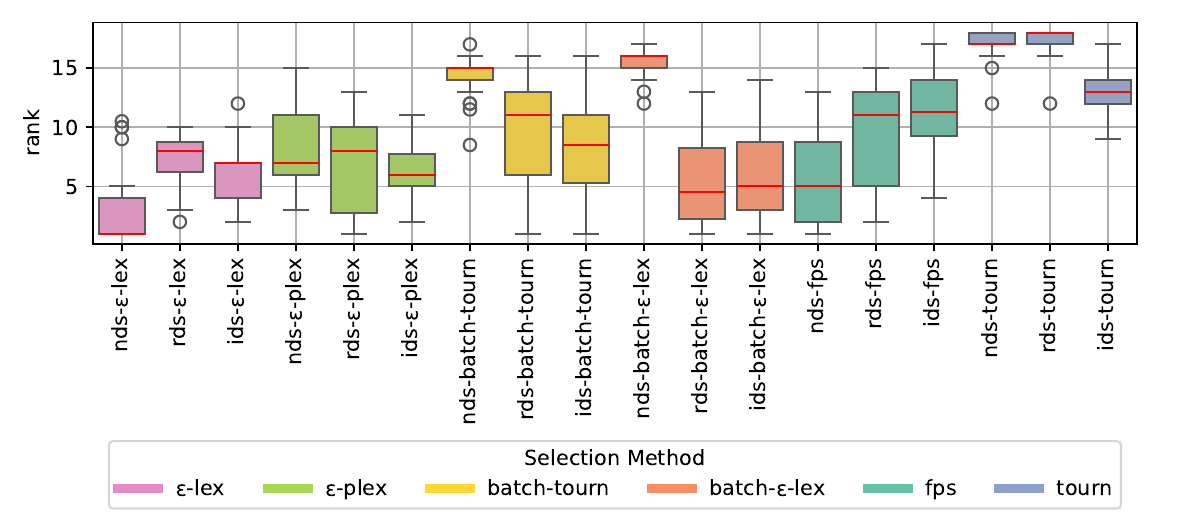}
    \caption{Ranking of the median \textbf{tree size} for all selection methods \textbf{given 1h}. The tree size is measured in terms of the median tree size of the final population (smaller is better).}
    \label{fig:rank_size_1h}
\end{figure}

\begin{figure}[h]
    \centering
    \includegraphics[scale=0.68]{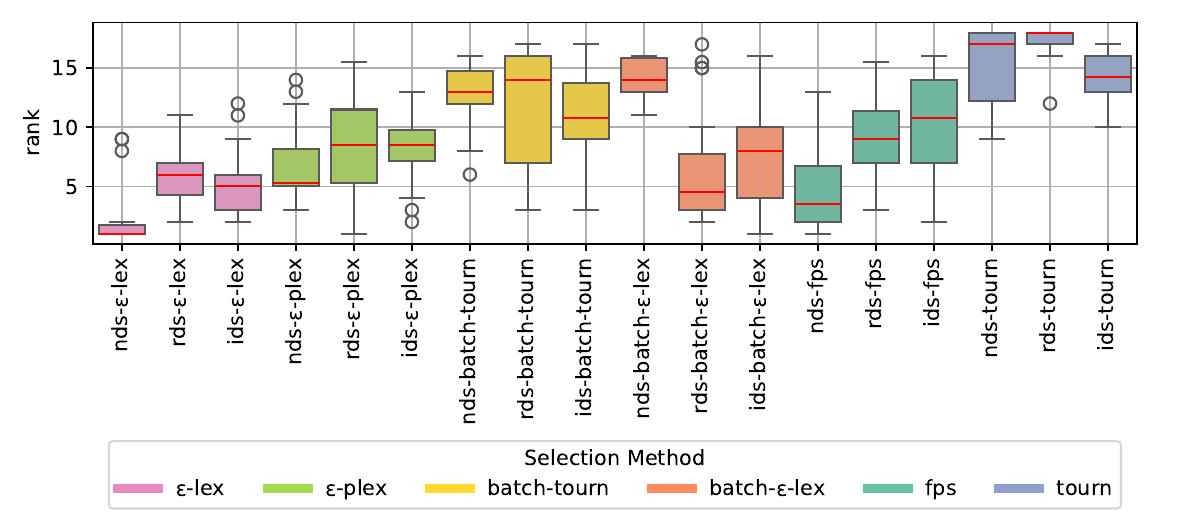}
    \caption{Ranking of the median \textbf{tree size} for all selection methods \textbf{given 15 minutes}. The tree size is measured in terms of the median tree size of the final population (smaller is better).}
    \label{fig:rank_size_15min}
\end{figure}

\clearpage
\section{Number of generations for a given running time}\label{appendix:gen}

\begin{table}[h] 
  \centering
    \caption{Median number of generations the search was performed for each method per problem given \textbf{24h}.} \label{tab:gen_24h}
 \begin{adjustbox}{width=1\textwidth}
\begin{tabular}{l|rrr|rrr|rrr|rrr|rrr|rrr}
\toprule
& \multicolumn{3}{c}{$\epsilon$-lex} & \multicolumn{3}{c}{$\epsilon$-plex} & \multicolumn{3}{c}{batch-tourn} & \multicolumn{3}{c}{batch-$\epsilon$-lex} & \multicolumn{3}{c}{fps} & \multicolumn{3}{c}{tourn}  \\
problem & nds & rds & ids & nds & rds & ids & nds & rds & ids & nds & rds & ids & nds & rds & ids & nds & rds & ids \\
\midrule
R-1027 & 726 & 7370 & 6103 & 8351 & 22476 & 20916 & 4545 & 17325 & 17384 & 3755 & 16578 & 15900 & 7193 & 12811 & 11063 & 1860 & 7780 & 12408 \\
R-1028 & 336 & 3402 & 3213 & 2474 & 14978 & 14358 & 1408 & 10892 & 10708 & 1350 & 11759 & 10461 & 2161 & 6757 & 6282 & 1462 & 6546 & 7406 \\
R-1030 & 406 & 3046 & 3330 & 2444 & 10330 & 17029 & 1267 & 9126 & 13438 & 1271 & 9630 & 11633 & 1464 & 4784 & 5670 & 1120 & 4574 & 9696 \\
R-207 & 2254 & 12898 & 13588 & 13544 & 60706 & 46500 & 9380 & 56998 & 44394 & 4073 & 35402 & 29886 & 10020 & 23680 & 20574 & 3148 & 15118 & 28002 \\
R-230 & 1606 & 10098 & 9642 & 7220 & 26814 & 27452 & 5908 & 40460 & 31076 & 3955 & 37480 & 22122 & 6164 & 14618 & 13192 & 2220 & 13750 & 18640 \\
R-4544 & 422 & 3105 & 2908 & 2123 & 10156 & 11688 & 1856 & 14958 & 7572 & 1190 & 18478 & 8328 & 4612 & 10143 & 4635 & 1062 & 3778 & 3706 \\
R-505 & 1466 & 8446 & 8775 & 8122 & 26980 & 27925 & 5440 & 22884 & 26122 & 4958 & 17550 & 18298 & 5307 & 13617 & 13404 & 4942 & 11057 & 15827 \\
R-519 & 1112 & 7856 & 7584 & 9913 & 27613 & 28882 & 8062 & 42878 & 31030 & 4604 & 27683 & 23184 & 5034 & 15894 & 16696 & 1717 & 15341 & 26258 \\
R-522 & 816 & 5908 & 6224 & 4828 & 25532 & 26411 & 2722 & 16230 & 25451 & 2266 & 19380 & 20948 & 3674 & 10538 & 9408 & 2044 & 9352 & 15754 \\
R-523 & 2723 & 15858 & 17674 & 21329 & 107828 & 57802 & 33116 & 91332 & 65650 & 11917 & 36316 & 33175 & 17044 & 47884 & 33117 & 8702 & 51382 & 51771 \\
R-560 & 1518 & 8370 & 7736 & 10196 & 35628 & 31050 & 5308 & 16673 & 20019 & 3048 & 19674 & 18777 & 7480 & 17854 & 15118 & 2672 & 10660 & 16252 \\
S-581 & 814 & 5880 & 5986 & 4096 & 20869 & 19360 & 4766 & 20992 & 25985 & 3700 & 23268 & 23724 & 4883 & 11162 & 10142 & 2191 & 5680 & 13904 \\
S-589 & 436 & 3294 & 3381 & 2275 & 10563 & 11888 & 2543 & 10180 & 12234 & 1870 & 10130 & 15216 & 3608 & 11379 & 6740 & 1401 & 5303 & 8656 \\
S-591 & 3288 & 18856 & 22648 & 14151 & 161357 & 146314 & 13128 & 89064 & 99588 & 7799 & 47478 & 47554 & 15996 & 28800 & 37904 & 7662 & 24487 & 53304 \\
S-606 & 441 & 3458 & 3312 & 2464 & 9402 & 9873 & 2643 & 10732 & 11333 & 1875 & 11988 & 13471 & 3260 & 11152 & 6688 & 1603 & 7839 & 8738 \\
S-607 & 430 & 3365 & 3076 & 2388 & 11134 & 11028 & 2627 & 11380 & 13130 & 2596 & 10345 & 12100 & 3035 & 7510 & 6181 & 1502 & 4510 & 7406 \\
S-615 & 1680 & 10123 & 10174 & 6908 & 40279 & 36341 & 7150 & 37958 & 46554 & 5534 & 32913 & 32580 & 8574 & 23412 & 15010 & 2092 & 8960 & 24454 \\
S-617 & 840 & 5614 & 6208 & 3800 & 19376 & 19294 & 4596 & 18181 & 19621 & 3164 & 18765 & 18404 & 3334 & 8770 & 10995 & 3220 & 10717 & 13954 \\
S-621 & 3148 & 18264 & 21908 & 14560 & 178452 & 156577 & 12414 & 82123 & 122315 & 6466 & 52349 & 54331 & 9024 & 24137 & 31875 & 4670 & 19896 & 50820 \\
S-623 & 442 & 3159 & 3008 & 2401 & 9734 & 9650 & 2214 & 5756 & 9557 & 1985 & 12600 & 10784 & 2235 & 6801 & 5496 & 1434 & 4269 & 7512 \\
S-624 & 3167 & 18160 & 20316 & 13643 & 161639 & 141082 & 11866 & 63680 & 109094 & 6092 & 47254 & 52357 & 9236 & 24248 & 32308 & 3826 & 17796 & 50134 \\
S-641 & 834 & 5664 & 6150 & 4000 & 21644 & 18472 & 3971 & 18260 & 24501 & 3240 & 19970 & 22982 & 3253 & 11094 & 11042 & 2242 & 7514 & 14272 \\
S-647 & 1538 & 9512 & 9960 & 7463 & 37023 & 36440 & 8894 & 40306 & 52378 & 5476 & 34008 & 35401 & 5430 & 14150 & 16736 & 3456 & 14160 & 27478 \\
S-654 & 819 & 5736 & 5986 & 3341 & 16753 & 19964 & 3598 & 19174 & 25880 & 1988 & 21656 & 24851 & 3512 & 10972 & 8972 & 1612 & 7434 & 15446 \\
R-665 & 2187 & 16156 & 14754 & 19353 & 82968 & 57665 & 8657 & 86554 & 55926 & 4325 & 38290 & 30011 & 14932 & 36544 & 23062 & 2896 & 16191 & 38984 \\
R-666 & 802 & 5207 & 5014 & 5016 & 15972 & 15552 & 2702 & 13404 & 14096 & 1942 & 18920 & 13328 & 3403 & 10388 & 8962 & 1378 & 5728 & 9424 \\
\bottomrule
\end{tabular}
\end{adjustbox}
\end{table}

\begin{table} 
    \centering
    \caption{Median number of generations the search was performed for each method per problem given \textbf{1h}.} \label{tab:gen_1h}
 \begin{adjustbox}{width=0.9\textwidth}
\begin{tabular}{l|rrr|rrr|rrr|rrr|rrr|rrr}
\toprule
& \multicolumn{3}{c}{$\epsilon$-lex} & \multicolumn{3}{c}{$\epsilon$-plex}  & \multicolumn{3}{c}{batch-tourn} & \multicolumn{3}{c}{batch-$\epsilon$-lex} & \multicolumn{3}{c}{fps} & \multicolumn{3}{c}{tourn} \\
problem & nds & rds & ids & nds & rds & ids & nds & rds & ids & nds & rds & ids & nds & rds & ids & nds & rds & ids \\
\midrule
R-1027 & 4 & 374 & 326 & 447 & 1361 & 1323 & 324 & 1013 & 942 & 269 & 831 & 910 & 1792 & 2467 & 2780 & 332 & 938 & 862 \\
R-1028 & 8 & 165 & 154 & 188 & 673 & 632 & 150 & 539 & 590 & 123 & 613 & 497 & 320 & 688 & 532 & 218 & 454 & 450 \\
R-1030 & 16 & 146 & 144 & 198 & 531 & 528 & 128 & 447 & 440 & 111 & 483 & 454 & 168 & 418 & 429 & 138 & 360 & 392 \\
R-207 & 117 & 608 & 584 & 1001 & 2958 & 2053 & 754 & 2546 & 2198 & 432 & 1376 & 1410 & 1042 & 2163 & 1558 & 468 & 1192 & 1740 \\
R-230 & 86 & 462 & 443 & 440 & 1332 & 1294 & 510 & 1874 & 1426 & 358 & 1556 & 951 & 675 & 1296 & 1146 & 400 & 1168 & 1078 \\
R-4544 & 17 & 160 & 146 & 262 & 806 & 653 & 174 & 896 & 464 & 148 & 1108 & 470 & 850 & 1124 & 649 & 236 & 411 & 436 \\
R-505 & 80 & 417 & 419 & 656 & 1488 & 1284 & 489 & 1184 & 1025 & 350 & 898 & 982 & 574 & 1170 & 914 & 492 & 852 & 1020 \\
R-519 & 49 & 341 & 336 & 457 & 1318 & 1422 & 402 & 1752 & 1374 & 266 & 1161 & 1068 & 364 & 1039 & 1106 & 234 & 871 & 1109 \\
R-522 & 38 & 284 & 278 & 358 & 1306 & 1358 & 242 & 944 & 990 & 206 & 837 & 946 & 343 & 940 & 812 & 270 & 771 & 842 \\
R-523 & 124 & 659 & 730 & 954 & 4719 & 2420 & 1212 & 4130 & 2628 & 462 & 1667 & 1227 & 884 & 2389 & 1614 & 758 & 2147 & 2216 \\
R-560 & 78 & 468 & 426 & 779 & 1586 & 1494 & 392 & 1078 & 1342 & 265 & 1032 & 1048 & 940 & 1770 & 1492 & 344 & 1034 & 1379 \\
S-581 & 46 & 298 & 294 & 409 & 1180 & 1458 & 395 & 1480 & 1627 & 325 & 1284 & 1266 & 439 & 1026 & 994 & 238 & 502 & 970 \\
S-589 & 27 & 184 & 181 & 242 & 815 & 753 & 258 & 714 & 898 & 240 & 690 & 876 & 304 & 744 & 572 & 176 & 496 & 590 \\
S-591 & 171 & 876 & 934 & 987 & 6670 & 6286 & 908 & 3980 & 4170 & 508 & 2067 & 2068 & 1080 & 2042 & 2292 & 640 & 1612 & 2752 \\
S-606 & 25 & 208 & 198 & 242 & 727 & 713 & 238 & 880 & 802 & 251 & 838 & 682 & 280 & 780 & 527 & 212 & 796 & 706 \\
S-607 & 26 & 172 & 159 & 257 & 801 & 742 & 230 & 848 & 794 & 189 & 812 & 769 & 306 & 812 & 900 & 180 & 412 & 650 \\
S-615 & 90 & 490 & 468 & 550 & 1996 & 1758 & 563 & 1866 & 2328 & 396 & 1715 & 1579 & 790 & 1706 & 1329 & 238 & 838 & 1527 \\
S-617 & 46 & 295 & 322 & 362 & 1116 & 1126 & 473 & 1250 & 1351 & 248 & 1027 & 1108 & 352 & 799 & 905 & 341 & 732 & 990 \\
S-621 & 166 & 880 & 904 & 932 & 7431 & 6614 & 864 & 4511 & 5210 & 502 & 2302 & 2266 & 777 & 1516 & 2098 & 508 & 1502 & 2914 \\
S-623 & 26 & 166 & 155 & 230 & 698 & 656 & 234 & 845 & 556 & 180 & 790 & 656 & 276 & 684 & 584 & 192 & 405 & 667 \\
S-624 & 166 & 846 & 852 & 884 & 6626 & 6115 & 740 & 3602 & 4761 & 444 & 2170 & 2252 & 704 & 1601 & 2093 & 446 & 1290 & 2866 \\
S-641 & 46 & 296 & 346 & 337 & 1229 & 1368 & 332 & 1185 & 1656 & 274 & 1045 & 1338 & 402 & 924 & 1036 & 248 & 641 & 1002 \\
S-647 & 85 & 468 & 458 & 542 & 1845 & 2044 & 657 & 2292 & 2645 & 378 & 1797 & 1590 & 654 & 1287 & 1184 & 388 & 1006 & 1610 \\
S-654 & 42 & 292 & 284 & 317 & 1100 & 1276 & 364 & 1274 & 1672 & 174 & 1211 & 1410 & 462 & 1002 & 864 & 228 & 622 & 1028 \\
R-665 & 114 & 684 & 619 & 1072 & 3648 & 2602 & 606 & 3144 & 1945 & 391 & 1641 & 1315 & 838 & 1774 & 1404 & 393 & 1394 & 1778 \\
R-666 & 34 & 262 & 247 & 354 & 938 & 908 & 228 & 648 & 672 & 176 & 813 & 654 & 564 & 1060 & 846 & 256 & 526 & 625 \\
\bottomrule
\end{tabular}
\end{adjustbox}
\end{table}

\begin{table} 
    \centering
    \caption{Median number of generations the search was performed for each method per problem given \textbf{15 minutes}.} \label{tab:gen_15min}
 \begin{adjustbox}{width=0.9\textwidth}
\begin{tabular}{l|rrr|rrr|rrr|rrr|rrr|rrr}
\toprule
 & \multicolumn{3}{c}{$\epsilon$-lex} & \multicolumn{3}{c}{$\epsilon$-plex} & \multicolumn{3}{c}{batch-tourn} & \multicolumn{3}{c}{batch-$\epsilon$-lex} & \multicolumn{3}{c}{fps} & \multicolumn{3}{c}{tourn} \\
problem & nds & rds & ids & nds & rds & ids & nds & rds & ids & nds & rds & ids & nds & rds & ids & nds & rds & ids \\
\midrule
R-1027 & 1 & 99 & 92 & 152 & 446 & 448 & 138 & 438 & 320 & 86 & 366 & 333 & 1348 & 2006 & 1860 & 221 & 524 & 525 \\
R-1028 & 2 & 44 & 43 & 74 & 212 & 195 & 68 & 224 & 177 & 47 & 216 & 150 & 248 & 362 & 284 & 102 & 198 & 166 \\
R-1030 & 2 & 42 & 40 & 74 & 175 & 159 & 54 & 154 & 127 & 42 & 124 & 138 & 82 & 168 & 168 & 59 & 138 & 138 \\
R-207 & 32 & 168 & 163 & 343 & 894 & 570 & 287 & 759 & 458 & 148 & 448 & 333 & 448 & 730 & 566 & 224 & 482 & 510 \\
R-230 & 24 & 132 & 127 & 169 & 403 & 374 & 179 & 582 & 420 & 122 & 404 & 292 & 292 & 526 & 449 & 190 & 456 & 364 \\
R-4544 & 2 & 46 & 40 & 102 & 298 & 201 & 80 & 332 & 184 & 54 & 288 & 192 & 423 & 578 & 324 & 142 & 197 & 179 \\
R-505 & 24 & 122 & 118 & 220 & 428 & 392 & 184 & 432 & 313 & 134 & 284 & 235 & 254 & 429 & 366 & 194 & 316 & 358 \\
R-519 & 10 & 92 & 88 & 138 & 361 & 366 & 118 & 442 & 382 & 90 & 341 & 250 & 138 & 362 & 352 & 112 & 290 & 321 \\
R-522 & 8 & 78 & 72 & 130 & 366 & 364 & 96 & 346 & 326 & 74 & 234 & 286 & 146 & 333 & 291 & 122 & 298 & 294 \\
R-523 & 33 & 169 & 188 & 294 & 1222 & 652 & 294 & 854 & 660 & 144 & 378 & 390 & 282 & 630 & 518 & 266 & 666 & 581 \\
R-560 & 19 & 126 & 120 & 247 & 492 & 432 & 158 & 476 & 418 & 100 & 292 & 343 & 462 & 744 & 608 & 178 & 398 & 428 \\
S-581 & 14 & 87 & 84 & 153 & 417 & 442 & 138 & 446 & 511 & 112 & 404 & 384 & 197 & 390 & 484 & 106 & 239 & 376 \\
S-589 & 6 & 54 & 54 & 102 & 280 & 242 & 99 & 256 & 220 & 80 & 224 & 228 & 144 & 282 & 226 & 80 & 203 & 197 \\
S-591 & 51 & 243 & 244 & 344 & 1758 & 1582 & 308 & 1042 & 1075 & 158 & 564 & 550 & 386 & 676 & 764 & 266 & 526 & 807 \\
S-606 & 6 & 57 & 54 & 96 & 266 & 232 & 98 & 310 & 217 & 90 & 242 & 198 & 120 & 284 & 205 & 84 & 306 & 246 \\
S-607 & 6 & 52 & 47 & 111 & 313 & 254 & 92 & 301 & 266 & 78 & 246 & 192 & 143 & 300 & 380 & 76 & 192 & 233 \\
S-615 & 26 & 138 & 126 & 222 & 588 & 542 & 194 & 603 & 707 & 134 & 505 & 434 & 372 & 602 & 558 & 122 & 328 & 470 \\
S-617 & 13 & 86 & 95 & 140 & 362 & 366 & 164 & 392 & 380 & 85 & 334 & 298 & 167 & 340 & 384 & 120 & 282 & 319 \\
S-621 & 48 & 232 & 232 & 304 & 1934 & 1676 & 300 & 1046 & 1294 & 158 & 603 & 591 & 322 & 536 & 710 & 218 & 538 & 879 \\
S-623 & 6 & 50 & 46 & 95 & 250 & 228 & 88 & 274 & 220 & 62 & 248 & 192 & 138 & 286 & 282 & 78 & 164 & 232 \\
S-624 & 48 & 228 & 218 & 306 & 1666 & 1552 & 246 & 918 & 1240 & 140 & 568 & 596 & 283 & 560 & 758 & 192 & 462 & 860 \\
S-641 & 14 & 87 & 104 & 138 & 431 & 420 & 138 & 376 & 451 & 90 & 307 & 324 & 186 & 361 & 455 & 108 & 266 & 400 \\
S-647 & 28 & 130 & 127 & 209 & 564 & 588 & 242 & 673 & 689 & 122 & 478 & 462 & 280 & 508 & 512 & 161 & 368 & 562 \\
S-654 & 12 & 81 & 82 & 126 & 365 & 393 & 148 & 382 & 400 & 70 & 358 & 335 & 260 & 424 & 380 & 104 & 242 & 357 \\
R-665 & 32 & 174 & 163 & 311 & 992 & 621 & 218 & 746 & 554 & 130 & 456 & 380 & 288 & 576 & 531 & 167 & 476 & 488 \\
R-666 & 6 & 73 & 70 & 128 & 312 & 298 & 106 & 317 & 214 & 64 & 228 & 192 & 307 & 516 & 392 & 130 & 240 & 236 \\
\bottomrule
\end{tabular}
\end{adjustbox}
\end{table}

\clearpage
\section{Results of the statistical analysis}\label{appendix:sig_test}

For our statistical analysis, we followed the recommendations of~\citeN{demvsar2006statistical} for comparing multiple methods over many datasets. For each configuration, we first use the Friedman test to test the null-hypothesis that all methods perform equivalent. We reject the null-hypothesis if $p < 0.05$. If the null-hypothesis is rejected, we use the Nemenyi post-hoc test for pairwise comparisons. The performance of two methods is significantly different if $p < 0.05$.  

\begin{figure}[h]
    \centering
    \includegraphics[scale=0.7]{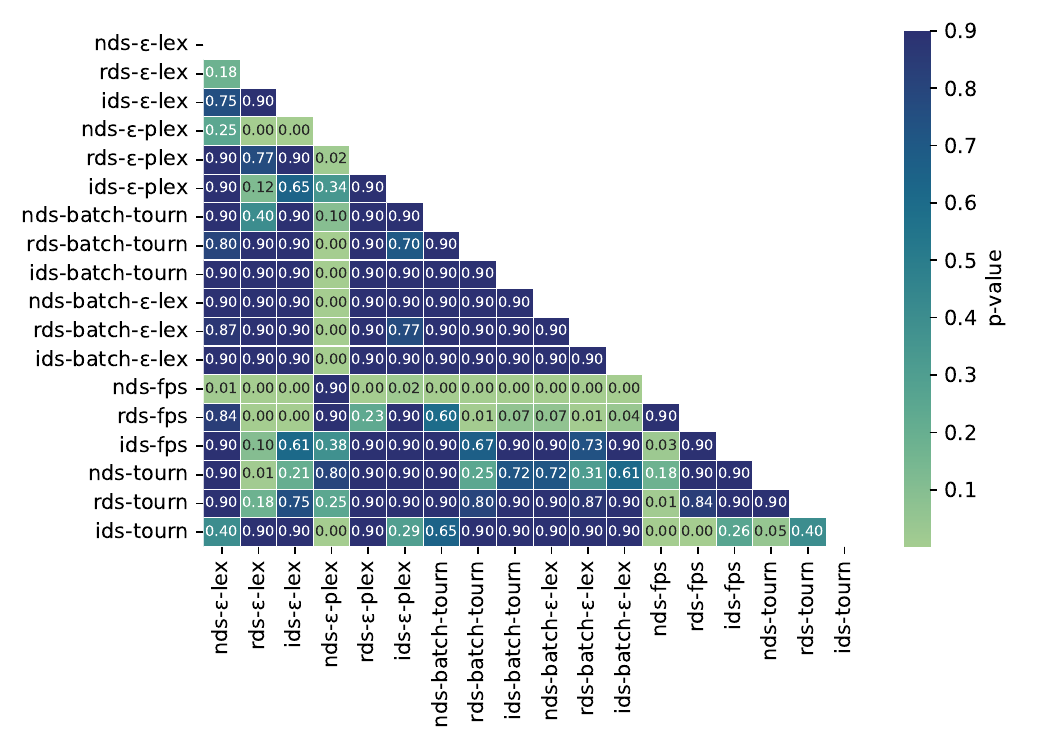}
    \caption{Results of the Nemenyi test for the results \textbf{given a fixed evaluation budget}. P-values are rounded to two decimal places.}
    \label{fig:sig_eval_budget}
\end{figure}

\begin{figure}[h]
    \centering
    \includegraphics[scale = 0.7]{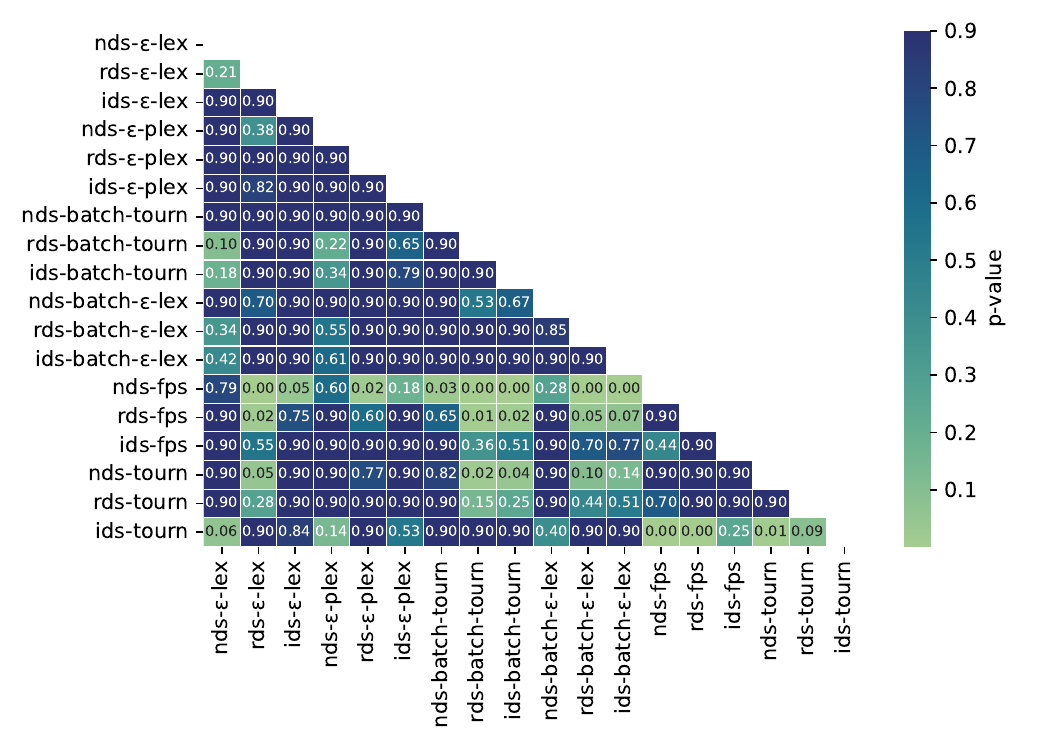}
    \caption{Results of the Nemenyi test for the results \textbf{given 24h}. P-values are rounded to two decimal places.}
    \label{fig:sig_24h}
\end{figure}

\begin{figure}[h]
    \centering
    \includegraphics[scale=0.7]{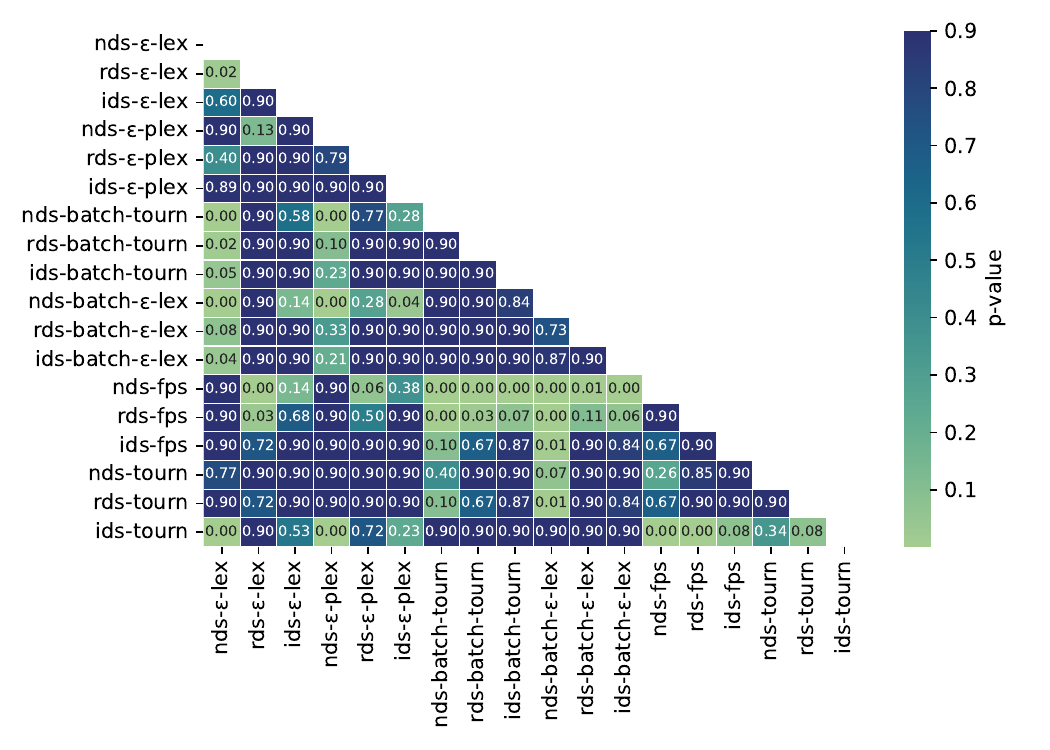}
    \caption{Results of the Nemenyi test for the results \textbf{given 1h}. P-values are rounded to two decimal places.}
    \label{fig:sig_1h}
\end{figure}

\begin{figure}[h]
    \centering
    \includegraphics[scale=0.7]{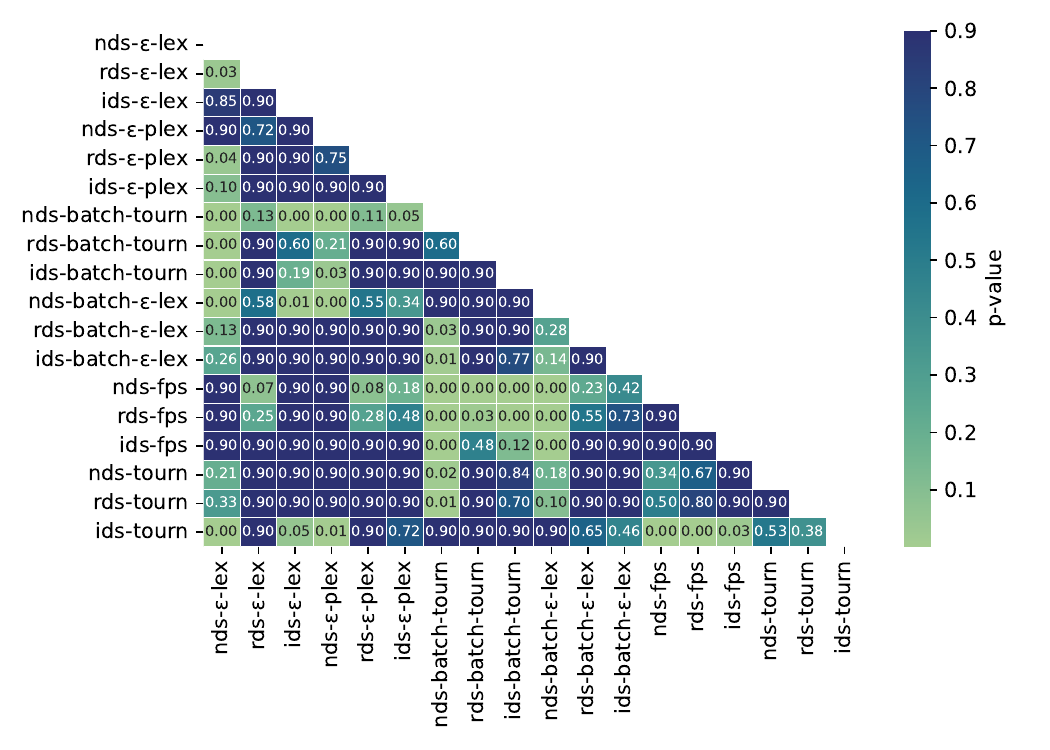}
    \caption{Results of the Nemenyi test for the results \textbf{given 15 minutes}. P-values are rounded to two decimal places.}
    \label{fig:sig_15min}
\end{figure}

The p-values of the Friedman test for each configuration are reported in Table~\ref{tab:friedman}. We reject the null-hypothesis in all cases. Therefore, we conducted the Nemenyi test for all configurations. Figures~\ref{fig:sig_eval_budget}-~\ref{fig:sig_15min} report the corresponding p-values.

\begin{table}[h]
    \centering
        \caption{Results of the Friedman test for all configurations.}
    \label{tab:friedman}
\begin{tabular}{lr}
\toprule
Configuration &  p-value \\
\midrule
Given a fixed evaluation budget &  3.92e-18 \\
Given 24h & 5.52e-11 \\
Given 1h & 1.37e-15 \\
Given 15 minutes &  1.03e-18 \\
\bottomrule
\end{tabular}
\end{table}

\end{document}